\documentclass{article}

\usepackage[main, final]{neurips_2026}
\setcitestyle{round,semicolon}

\usepackage[utf8]{inputenc}
\usepackage[T1]{fontenc}
\usepackage{amsmath}
\usepackage{hyperref}
\usepackage{url}
\usepackage{booktabs}
\usepackage{amsfonts}
\usepackage{amsthm}

\usepackage{nicefrac}
\usepackage{microtype}
\usepackage{graphicx}
\usepackage[dvipsnames,table]{xcolor}
\usepackage{subcaption}
\usepackage{multirow}
\usepackage{wrapfig}
\definecolor{darkred}{RGB}{139,0,0}
\definecolor{darkgray}{RGB}{80,80,80}
\definecolor{darkgreen}{RGB}{0,100,0}
\usepackage{tcolorbox}
\usepackage{makecell}
\usepackage{etoc}
\usepackage{todonotes}
\usepackage{array}

\definecolor{tticblue}{RGB}{0, 94, 184}

\newif\ifDRAreviewnotes
\DRAreviewnotestrue
\AtBeginDocument{\setlength{\marginparwidth}{82pt}}

\let\cite\citep

\title{Expanding the Role of Diffusion Models for Robust Classifier Training}

\author{%
  Pin-Han Huang \\
  National Taiwan University \\
  \texttt{pinhankhuang@gmail.com}
  \And
  Shang-Tse Chen\footnotemark[2] \\
  National Taiwan University \\
  \texttt{stchen@csie.ntu.edu.tw}
  \And
  Hsuan-Tien Lin\footnotemark[2] \\
  National Taiwan University \\
  \texttt{htlin@csie.ntu.edu.tw}
}

\begin{document}

\maketitle
\begingroup
\renewcommand{\thefootnote}{\fnsymbol{footnote}}%
\footnotetext[2]{Co-advisors.}%
\endgroup
\etocdepthtag.toc{mtchapter}

\begin{abstract}
Incorporating diffusion-generated synthetic data into adversarial training (AT) has been shown to substantially improve the training of robust image classifiers.
In this work, we extend the role of diffusion models beyond merely generating synthetic data, examining whether their internal representations, which encode meaningful features of the data, can provide additional benefits for robust classifier training.
Through systematic experiments, we show that diffusion models offer representations that are both diverse and partially robust, and that explicitly incorporating diffusion representations as an auxiliary learning signal during AT consistently improves robustness across settings.
Furthermore, our representation analysis indicates that incorporating diffusion models into AT encourages more disentangled features, while diffusion representations and diffusion-generated synthetic data play complementary roles in shaping representations.
Experiments on CIFAR-10, CIFAR-100, and ImageNet validate these findings, demonstrating the effectiveness of jointly leveraging diffusion representations and synthetic data within AT.

\end{abstract}

\section{Introduction}

Machine learning models are known to be vulnerable to adversarial examples~\cite{szegedy2014intriguing, goodfellow2015explaining}, inputs perturbed by semantically imperceptible noise that can drastically alter model predictions. Among numerous proposed defenses, adversarial training (AT)~\cite{madry2018towards, zhang2019trades}, which trains models on adversarially perturbed inputs,
remains one of the most effective approaches on standard benchmarks such as RobustBench~\cite{croce2021robustbench}.

Despite its effectiveness, AT is known to suffer from robust overfitting~\cite{rice2020overfitting}, where robustness on the test set degrades during training despite decreasing training loss and stable clean accuracy. Multiple methods have been proposed to understand and mitigate this issue~\cite{wu2020awp, chen2021robust, yu2022understand, wang2023balance, wang2023better, wu2024annealing}. Among them, arguably the most effective approach to date has been the diffusion model with AT (DM-AT) training recipe~\cite{wang2023better}, leveraging large amounts of high-quality synthetic data generated by diffusion models to improve robustness.

The DM-AT approach \citep{wang2023better}, which does not rely on additional real data to train diffusion models, mainly treats diffusion models as synthetic data generators for robust classifier training. More broadly, most of the existing efforts to improve AT have centered on this synthetic data paradigm~\cite{ouyang2023improving, bartoldson2024adversarial,cui2024dkl, wu2025vision}. While effective, this paradigm largely exploits diffusion models only at the level of their outputs, potentially overlooking the rich structure developed during the generative process. In particular, diffusion models are known to produce meaningful intermediate representations throughout their denoising trajectories~\citep{yang2023diffusion, xiang2023ddae, chen2025deconstructing, li2025understanding}. The extent to which these representations can further enhance robustness remains largely unexplored, suggesting a promising opportunity beyond synthetic data generation.
In this work, we systematically investigate the utility of diffusion representations for enhancing adversarial robustness. We hypothesize that the denoising objective encourages diffusion models to encode resilient semantic features from corrupted inputs, which can be harnessed to guide classifier training. Specifically, we examine the utility of intermediate activations computed during denoising as feature priors for improving AT.

\begin{wrapfigure}{r}{0.40\textwidth}
    \vspace{-0.4cm}
    \centering
    \includegraphics[width=\linewidth]{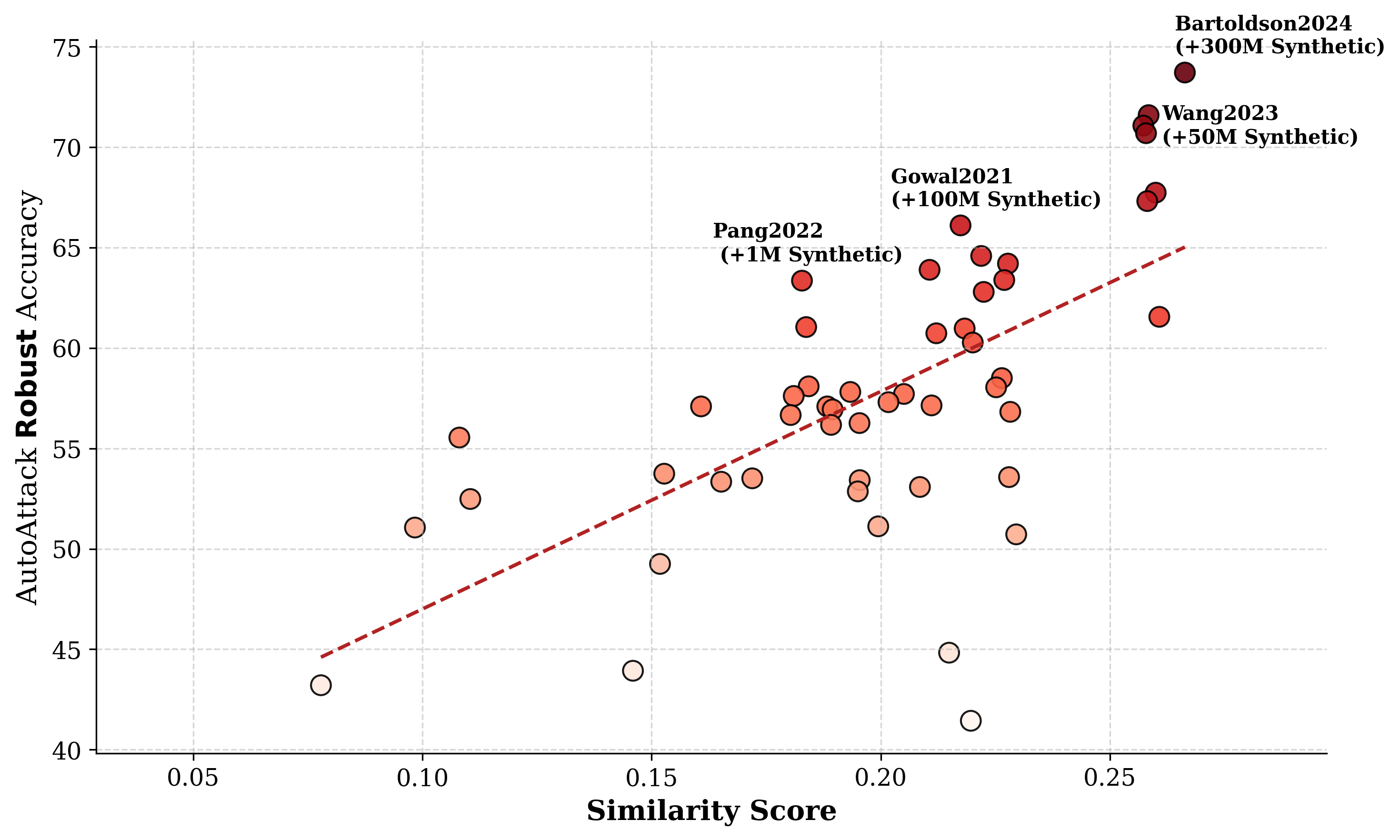}
    \vspace{-0.6cm}
    \caption{Robust accuracy versus similarity to diffusion representations for CIFAR-10 $\ell_\infty$-robust models from RobustBench. Details are in Appendix~\ref{app:imp_alignment_trend}.}
    \label{fig:align_fig_1}
    \vspace{-0.4cm}
\end{wrapfigure}

Our idea of studying the intermediate activations as feature priors is inspired by an intriguing analysis that uncovers a weak correlation between robustness and alignment with diffusion representations (Figure~\ref{fig:align_fig_1}). 
Going further, we find that the extracted diffusion representations possess a range of highly desirable properties: they capture rich and diverse information, emphasize lower-frequency structures, and exhibit strong resilience to irrelevant high-frequency noise.
These characteristics suggest that diffusion representations have significant potential, in contrast to conventional reconstruction-based representation learning, which is known to be more adversarially vulnerable due to its reliance on high-frequency signals~\cite{huang2023improving}.

Motivated by these findings, we propose to upgrade the DM-AT recipe by incorporating a simple module that aligns classifier representations with diffusion representations~\cite{li2023dream, yu2025repa, stracke2025cleandift}.
The modified recipe uses diffusion representations as an auxiliary learning signal to guide feature learning during robust classifier training, and is agnostic to the choice of classifier backbone. Extensive experiments on CIFAR-10, CIFAR-100, and ImageNet across multiple architectures and diffusion synthetic data settings demonstrate consistent improvements in both clean and robust accuracy,
showing the effectiveness of using the semantic features encoded in diffusion representations to guide feature learning during AT.

We deepen our analysis to understand the mechanism behind the empirical improvements by building on the recent framework of mechanistic interpretability~\cite{gorton2025advsuper}, comparing the effects of synthetic data in DM-AT versus the diffusion representation alignment in our upgraded recipe. This approach reveals that both interventions facilitate the learning of more easily disentangled representations, yet they achieve this effect through distinct underlying mechanisms. 
Specifically, our analysis, guided by classification dimension~\cite{feng2022rank}, reveals that synthetic data in DM-AT promotes robustness and generalization by enabling the model to learn low-rank representations with strong generalization properties. 

In contrast, diffusion representation alignment encourages the model to effectively leverage its representational dimensions to encode robust features, which are not necessarily low-rank. Together, these findings suggest that our proposed diffusion representation alignment complements the synthetic data used in DM-AT for robust classifier training, supporting our goal of \emph{expanding} the role of diffusion models in robustness and generalization.

Our contributions are summarized as follows:
\begin{itemize}
    \item We show that diffusion representations encode features that are partially robust and diverse, and leveraging diffusion representations as an auxiliary learning signal improves AT.
    \item We find that the incorporation of diffusion models encourages models to learn representations that are easier to disentangle, with synthetic data and representation alignment playing complementary roles.
    \item Extensive experiments on CIFAR-10, CIFAR-100, and ImageNet show that incorporating both diffusion representation alignment and diffusion synthetic data consistently improves robustness, offering an updated recipe to build robust classifiers.
\end{itemize}

\section{Related Work}
\textbf{Adversarial Robustness.} Adversarial robustness studies the stability of model predictions under adversarially crafted imperceptible perturbations~\cite{szegedy2014intriguing,goodfellow2015explaining}. 
Empirical robustness is commonly assessed with AutoAttack~\cite{croce2020reliable}, which is also the main evaluation protocol for RobustBench~\cite{croce2021robustbench}. 
RobustBench excludes defenses that rely on inference-time randomness or optimization loops, since such mechanisms are frequently broken by adaptive attacks and require more careful and costly evaluations~\cite{athalye2018obfuscated, gao2022LimitPreprocess}. Consequently, models competitive on RobustBench rely on adversarial training to achieve robustness. Additionally, certified defenses offer provable guarantees~\cite{cohen2019RS, carlini2023certified, hu2024a, chen2024dmcrc, lai2025enhancing}, but often introduce significant inference-time cost or underperform adversarially trained models in empirical robustness, especially models that are trained with diffusion synthetic data~\cite{wang2023better, bartoldson2024adversarial, wu2025vision}. In this work, we focus on improving adversarial training and adopt AutoAttack as our primary evaluation.

\textbf{Diffusion Representations.} Diffusion models have achieved remarkable success in image generation~\cite{Ho2020ddpm, dhariwal2021dmbeatsgan, song2021scorebased, rombach2022ldm, karras2022edm, ma2024SiT, yu2025repa, yao2025ldit}, and studies in representation learning have shown that intermediate activations extracted from diffusion models are effective for discriminative tasks, including competitive performance compared with other self-supervised learning methods for image classification~\cite{xiang2023ddae, chen2025deconstructing, li2025understanding} and dense predictions~\cite{stracke2025cleandift, gan2025unleashing}. The latest advances in diffusion models have also leveraged this insight to accelerate the training of diffusion models by aligning the activations with large-scale self-supervised learning encoders~\cite{yu2025repa, singh2026what}.

In this paper, we investigate whether diffusion representations encode informative and robust semantics that can benefit adversarially robust classification.
\citet{yagoda2025diffrobust} propose to train prediction heads on frozen unconditional diffusion representations as a lightweight approach, but it relies heavily on inference-time randomness and is less robust than adversarial training.
Under EOT-based evaluation~\cite{athalye2018obfuscated}, the robust accuracy drops substantially (Appendix~\ref{app:det_for_related_work}).
Conversely, we show that using diffusion representations as an auxiliary learning signal can further strengthen adversarial training, and analyze how this integration shapes the learned representations.

\textbf{Diffusion Purification and Generative Classifiers.} In addition to generating synthetic data for adversarial training, diffusion models have also been applied in adversarial purification to remove adversarial noise~\cite{nie2022diffpure, li2025adbm}. However, such methods incur substantial inference cost, and their reliance on randomness has been shown to be vulnerable to adaptive attacks~\cite{wang2024diffhammer, chen2024rcsdm}. 

Another direction is to turn off-the-shelf diffusion models into Bayesian generative classifiers~\cite{li2023yourdiffusion, clark2023t2i, chen2024rcsdm}.
At a high level, these methods add noise to an input image, then denoise it conditioned on each class, and finally select the class whose reconstruction is most similar to the original input.
They exhibit desirable properties such as high error consistency with humans~\cite{jaini2024intriguing}, robustness to spurious correlations~\cite{li2025generative} that is free of the need for a balanced dataset~\cite{kirichenko2023last}, and adversarial robustness~\cite{chen2024rcsdm}.
These approaches can also be integrated with randomized smoothing~\cite{cohen2019RS} to provide certified defenses~\cite{chen2024dmcrc}.
Despite these advantages, the approach incurs substantial inference overhead due to iterative denoising and class-conditional evaluation, which scales inference cost with the number of classes and limits practicality for deployment and full evaluation on datasets such as ImageNet~\cite{li2023yourdiffusion, chen2024rcsdm}.
In this work, we pursue a parallel path that focuses on leveraging diffusion models to enhance robust classifier training, which is free of inference-time overhead.

\section{Preliminaries}
\textbf{Adversarial Training (AT).} 
Given a dataset $\mathcal{D}=\{(\mathbf{x}_i, y_i)\}_{i=1}^n$ of image-label pairs, adversarial training~\cite{madry2018towards} is formulated as
\begin{equation}
\min_{\theta}\sum_{i=1}^{n}\max_{\mathbf{x}_i' \in \mathcal{S}(\mathbf{x}_i)} 
\mathcal{L}\!\big(f_{\theta}(\mathbf{x}_i'),\, y_i\big),
\end{equation}
where $f_{\theta}$ is the model parameterized by $\theta$, $\mathcal{L}$ is the cross-entropy loss, and
$\mathcal{S}(\mathbf{x})=\left\{\mathbf{x}' : \lVert \mathbf{x}'-\mathbf{x}\rVert_p \le \varepsilon \right\}$ is the $\ell_p$-ball of radius $\varepsilon$ centered at $\mathbf{x}$. During training, projected gradient descent (PGD) is used to approximately solve the inner maximization by iteratively updating the adversarial example. 
Considering the standard $\ell_\infty$ adversary, the adversarial example is obtained by
\begin{equation}
\mathbf{x}_i^{(t+1)}=\Pi_{\mathcal{S}(\mathbf{x}_i)}\!\left(\mathbf{x}_i^{(t)}+\alpha\,\mathrm{sign}\!\left(\nabla_{\mathbf{x}}\mathcal{L}\big(f_{\theta}(\mathbf{x}_i^{(t)}),y_i\big)\right)\right),
\end{equation}
where $\alpha$ is the step size and $\Pi_{\mathcal{S}(\mathbf{x}_i)}(\cdot)$ denotes projection onto $\mathcal{S}(\mathbf{x}_i)$ and the valid image pixel range.
The TRADES loss~\citep{zhang2019trades} is often used to balance the performance on adversarial and clean images~\citep{wang2023better, bartoldson2024adversarial, cui2024dkl, wu2025vision}.

\label{sec:prelim_extract_rep}
\textbf{Extracting Diffusion Representations.} For a diffusion model $g_\phi$, it can be seen to be composed of an encoder $g_{\phi_{\text{enc}}}$ and a decoder $g_{\phi_{\text{dec}}}$.
Given a denoising timestep $t$, the corresponding noisy image $\mathbf{x}_t$, and optional conditions $\mathbf{c}$, we refer to the output of the encoder, $g_{\phi_{\text{enc}}}(\mathbf{x}_t, t, \mathbf{c})=\mathbf{h}^{\text{DR}}_{\mathbf{x}_t,t,\mathbf{c}}$, as diffusion representations. In practice, for UNet-based diffusion models, representations are typically extracted from the upsampling blocks near the bottleneck layer~\cite{xiang2023ddae}, whereas for newer transformer-based diffusion models~\cite{peebles2023dit}, representations are extracted near the middle layers~\cite{xiang2023ddae, chen2025deconstructing}. Additionally, the representation quality of diffusion models is often unimodal across timesteps, peaking at timesteps where the noisy image $\mathbf{x}_t$ contains a small amount of noise that removes irrelevant details for the perception task.
This behavior can be explained by the high signal-to-noise ratio within the class-relevant subspace at those timesteps~\cite{li2025understanding}.
In this work, we extract the representations from timesteps based on linearly probed clean accuracy experiments following prior work~\cite{xiang2023ddae}.
Implementation details are provided in Appendix~\ref{app:DRA_Imp_Abl}.

\section{Methodology}
\begin{figure*}[t]
  \centering
  \begin{subfigure}[t]{0.70\textwidth}
    \centering
    \includegraphics[width=\linewidth, height=4.2cm, keepaspectratio]{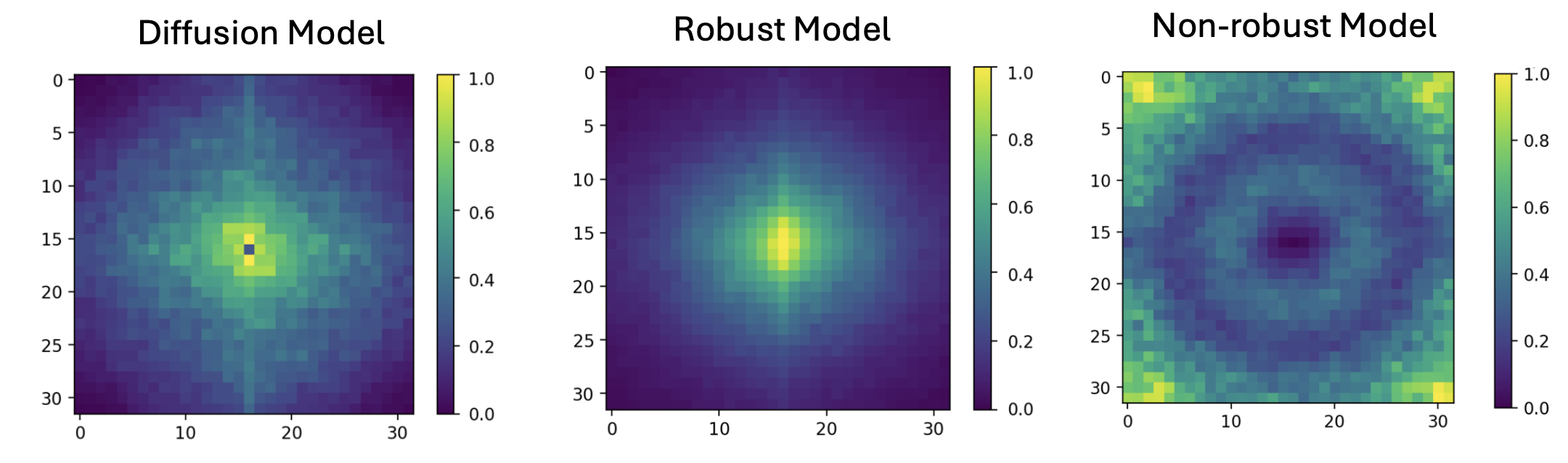}
    \caption{}
    \label{fig:freq_map}
  \end{subfigure}\hfill
  \begin{subfigure}[t]{0.30\textwidth}
    \centering
    \includegraphics[width=\linewidth, height=4.2cm, keepaspectratio]{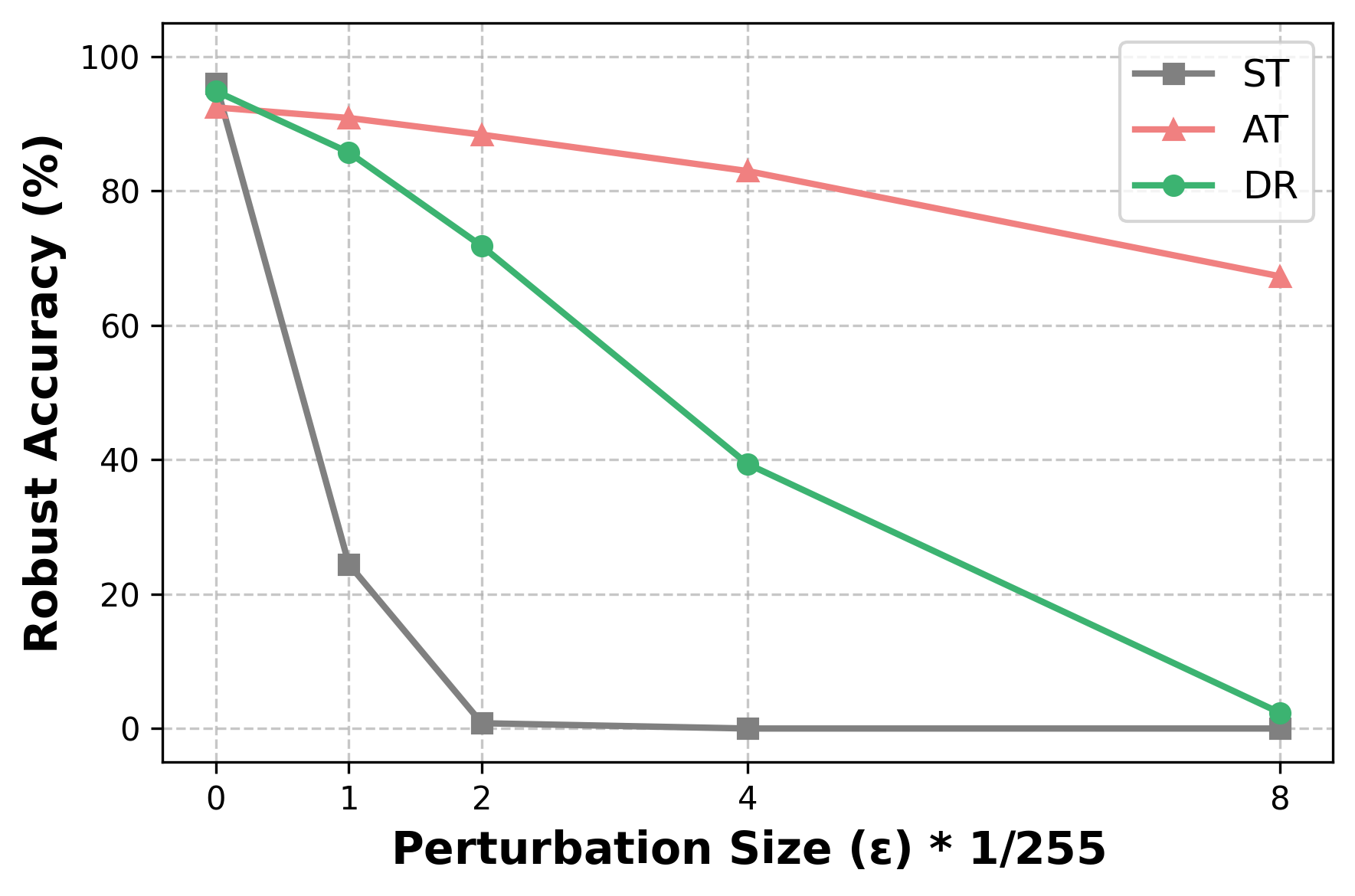}
    \caption{}
    \label{fig:robust_acc}
  \end{subfigure}
  \caption[Frequency saliency and robust accuracy of diffusion representations.]{(a) The frequency saliency of the linearly probed diffusion representation, adversarially trained robust model, and standard non-robust model. Low frequencies are centered. (b) The CIFAR-10 robust accuracy across perturbation budgets for the linearly probed diffusion representation (\textcolor{darkgreen}{DR}), adversarially trained robust model (\textcolor{darkred}{AT}), and standard-trained non-robust model (\textcolor{darkgray}{ST}).}
  \label{fig:sec4_properties}
  \vspace{-0.3cm}
\end{figure*}

Prior work suggests diffusion models may be less effective for representation learning and that pixel-reconstruction objectives can encourage non-informative or high-frequency features that hurt downstream adversarial training~\cite{yu2025repa,balestriero2024how,huang2023improving}.
In this section, we empirically show that diffusion models trained via noisy-image denoising in fact learn features with desirable properties. Building on this, we leverage diffusion features as an auxiliary learning signal to improve downstream adversarial training. Additional discussion and analysis based on representation similarity are provided in Appendix~\ref{app:imp_alignment_trend}.

\subsection{Observation}

\begin{wrapfigure}{r}{0.42\textwidth}
    \vspace{-0.5cm}
    \centering
    \includegraphics[width=\linewidth]{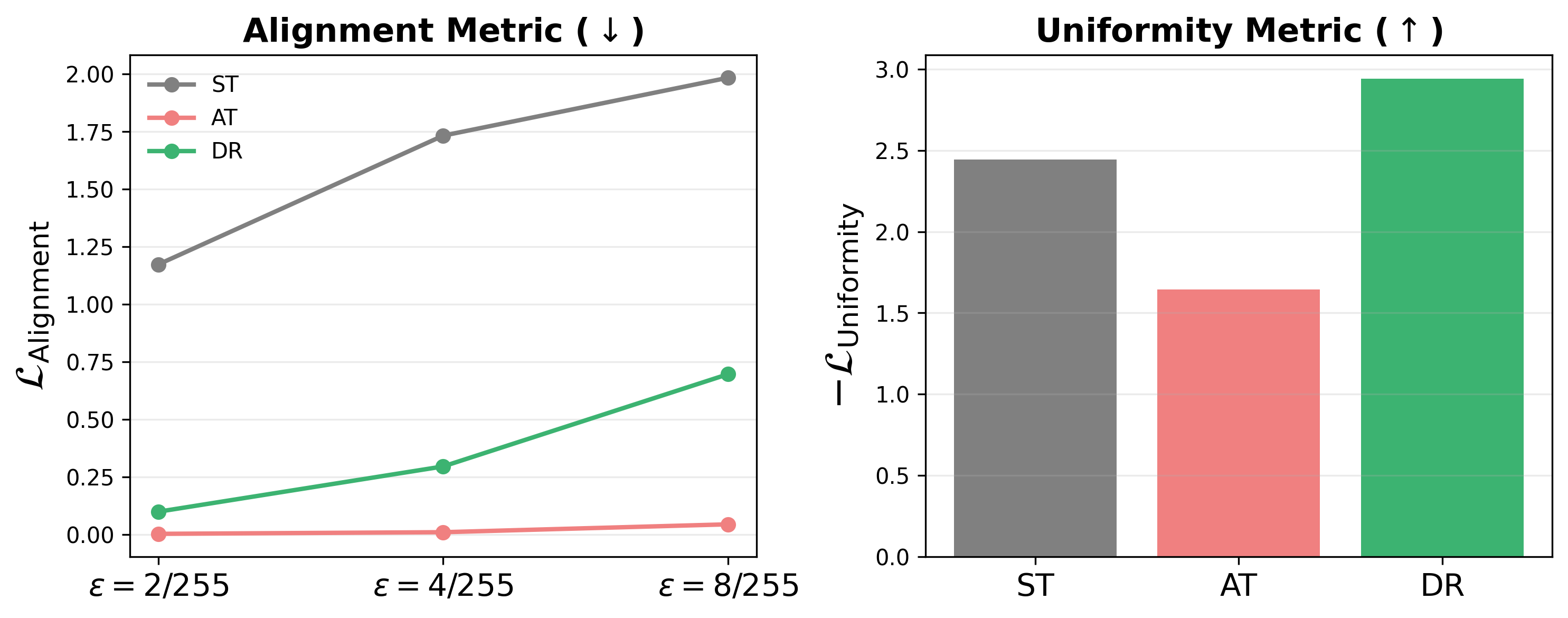}
    \vspace{-0.6cm}
    \caption{Representation metrics on CIFAR-10 for the standard-trained model (\textcolor{darkgray}{ST}), the adversarially trained model (\textcolor{darkred}{AT}), and the diffusion representations (\textcolor{darkgreen}{DR}).}
    \label{fig:rep_metric}
    \vspace{-0.5cm}
\end{wrapfigure}

\textbf{Representation Metrics.} We posit that diffusion models encode representations that are inherently mildly robust but preserve diversity that can improve downstream adversarial training.
To investigate the hypothesis, we analyze these representations using two key metrics from the representation learning literature: uniformity and alignment~\cite{wang2020understanding}.
Uniformity measures how evenly representations are distributed on the unit hypersphere, reflecting the information preserved in the representation space,  whereas alignment measures the distance between the representations of data examples with different positive views. In our setting, we form positive pairs by constructing adversarial images.
As shown in Figure~\ref{fig:rep_metric}, diffusion representations are more robust than standard supervised training, while achieving richer features with a noticeably higher uniformity metric.
In contrast, adversarial training increases alignment but decreases feature diversity, reflecting the difficulty of learning a robust model that preserves feature diversity.
In this work, we aim to leverage diffusion representations to improve adversarial training by shifting the alignment--uniformity frontier.

\textbf{Frequency and Robustness Behavior.}
Vision models are often sensitive to high-frequency input perturbations, while adversarial training typically reduces this sensitivity and places more emphasis on low-frequency components~\cite{chan2022how}.
Moreover, pixel reconstruction-based pretraining like MAE has been reported to have worse adversarially trained downstream performance than standard supervised training, explained by a stronger reliance on mid and high-frequency features~\cite{huang2023improving}.
To investigate if diffusion representations exhibit similar behaviors, we conduct frequency-saliency analysis on the PGD perturbations on the CIFAR-10 linearly probed diffusion representations.
Figure~\ref{fig:freq_map} shows that diffusion representations exhibit lower high-frequency saliency that resembles robust models.
It also suggests that diffusion representations do not suffer from the same pixel-reconstruction frequency behavior reported in previous work, which is likely related to the partially noise-corrupted images seen during denoising training.

Additionally, one possibility is that frozen diffusion features are already robust enough and do not require further robust finetuning~\cite{yagoda2025diffrobust}.
We evaluate robustness by measuring the robust accuracy of a linear probe trained on top of a frozen unconditional diffusion model.
To avoid a false sense of robustness due to randomness \cite{athalye2018obfuscated}, we make the noise used during diffusion feature extraction deterministic. Figure~\ref{fig:robust_acc} shows that these representations are inherently more robust to small-budget perturbations than standard supervised finetuned models, but still require robust training for competitive robustness.

\begin{wrapfigure}{r}{0.42\textwidth}
    \vspace{-0.3cm}
    \centering
    \includegraphics[width=\linewidth]{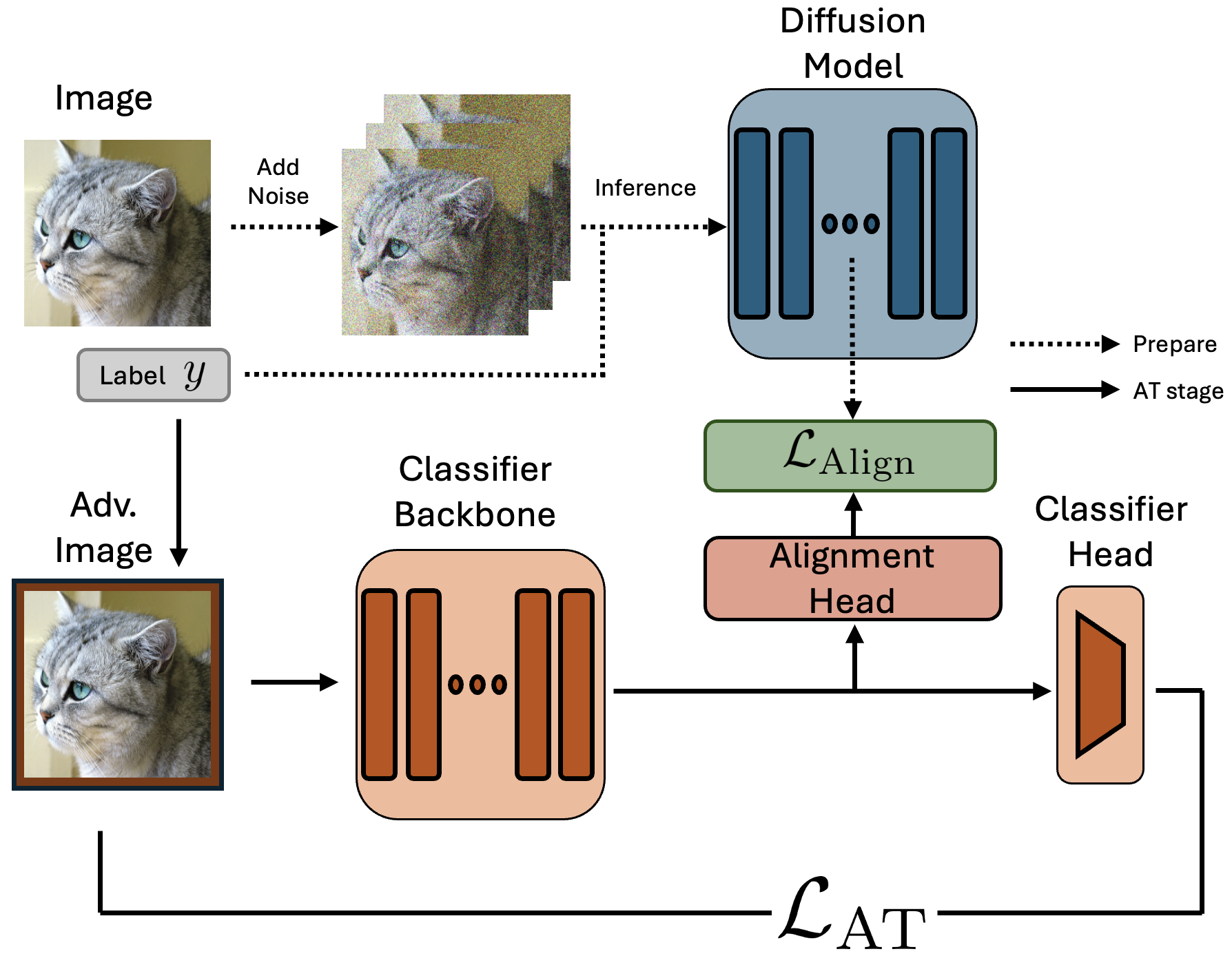}
    \vspace{-0.6cm}
    \caption{Diffusion Representation Alignment (DRA). We leverage an auxiliary projection head to align classifiers with the extracted diffusion representations.}
    \label{fig:method_overview}
\end{wrapfigure}

\subsection{Diffusion Representation Alignment for Robust Classifier Training}

\label{sec:DRA}

To efficiently improve downstream adversarial training with diffusion representations, we propose to integrate adversarial training with representation alignment~\cite{yang2023diffusion, yu2025repa, stracke2025cleandift}.
Figure~\ref{fig:method_overview} illustrates the overall framework, where we leverage an auxiliary projection head to regularize the classifier representations.

Given an image-label pair $(\mathbf{x}, y)$, and the classifier $f_{\text{CLS}}=g_{\theta} \circ f_{\theta}$ consisting of an encoder $f_{\theta}$ and a classification head $g_{\theta}$,
we regularize the classifier representation $\mathbf{h}^{\text{CLS}}_{\hat{\mathbf{x}}}=f_{\theta}(\hat{\mathbf{x}})$ given the adversarial example $\hat{\mathbf{x}}$ computed during training. Specifically, we align classifier representations with the representations
$\mathbf{h}^{\text{DR}}_{\mathbf{x}_t,t,y}$
extracted from a frozen diffusion model (Section~\ref{sec:prelim_extract_rep}), using a trainable projection head $g_{\text{proj}}$ that maps between the representation spaces. The diffusion representation alignment loss is defined as
\begin{equation}
\mathcal{L}_{\text{DRA}}
= - \mathrm{sim}(
g_{\text{proj}}(\mathbf{h}^{\text{CLS}}_{\hat{\mathbf{x}}}),
\mathbf{h}^{\text{DR}}_{\mathbf{x}_t,t,y}),
\end{equation}
where $\mathrm{sim}(\cdot,\cdot)$ is the given representation similarity metric.
The overall training objective becomes
\begin{equation}
\mathcal{L}_{\text{AT-DRA}}
= \mathcal{L}_{\text{AT}} + \lambda \mathcal{L}_{\text{DRA}},
\end{equation}
where $\lambda$ controls the regularization strength.
In practice, we implement the projection head $g_{\text{proj}}$ with an MLP module and use cosine similarity as the similarity metric, which performs well empirically and matches the prior work recommendations~\citep{yu2025repa, stracke2025cleandift}.
For each image, we use a fixed timestep around the optimal linearly probed timestep~\citep{xiang2023ddae} to extract diffusion representations.
Details on extracting diffusion representations, projection head, and timestep choices are in Appendix~\ref{app:DRA_Imp_Abl}.

\section{Experiments}
In this section, we evaluate the effectiveness of Diffusion Representation Alignment (DRA) with adversarial training.
The experiments across CIFAR-10, CIFAR-100, and ImageNet with different architectures and settings show consistent improvements.
Lastly, we analyze the effect on classifier representations when incorporating diffusion models into adversarial training.

\subsection{Setup}
\label{sec:exp_setups}
We mainly follow the DM-AT~\cite{wang2023better} setup, which is the state-of-the-art adversarial training framework that is also used by~\citet{bartoldson2024adversarial, cui2024dkl, wu2025vision}.
In the following, we briefly explain the setup for each dataset. Implementation details are in Appendix~\ref{app:exp_setup_details}.

\textbf{CIFAR-10/100.} For CIFAR-10/100~\cite{krizhevsky2009learning}, we follow the DM-AT recipe and perform AT with $\ell_\infty$ perturbation budget $\varepsilon=8/255$ and step size $\alpha=2/255$, using 10 PGD steps to adversarially augment the training images.
The TRADES loss~\cite{zhang2019trades} is used as the adversarial training objective.
To demonstrate the effectiveness of the proposed method across different training budget setups, CIFAR-10 experiments are conducted with synthetic data containing 1 million, 20 million, and 50 million synthetic images released by~\citet{wang2023better}, with DRA using the same frozen CIFAR-10 EDM diffusion model checkpoint~\cite{karras2022edm} that was used to generate the synthetic images.
Additionally, experiments with CIFAR-100 using 1 million synthetic images and the EDM model released by \citet{wang2023better} are also provided. 
For model choices, we conduct experiments on the WideResNets~\cite{zagoruyko2016wrn} WRN-28-10 and WRN-34-10, which are widely used in the adversarial training literature, as well as ViT-B/2~\cite{dosovitskiy2021vit} following~\citet{wu2025vision}.

\textbf{ImageNet.} Additionally, to demonstrate the effectiveness in real-world scenarios, we conduct experiments on ImageNet~\cite{russakovsky2015imagenet}, with models initialized from strong pretrained checkpoints.
We set the perturbation budget to $\varepsilon=4/255$ and evaluate with an input resolution of $224 \times 224$ following the RobustBench standard.
For model selection, we train and evaluate ViT-B/16 and ConvNeXt-B~\cite{liu2022convnext}, initialized from the DINOv3~\cite{simeoni2025dinov3} pretrained checkpoints released via \texttt{timm}~\cite{rw2019timm}.
For the diffusion synthetic data, we generate 4 million $256 \times 256$ synthetic images using LightningDiT~\cite{yao2025ldit}.

\subsection{Diffusion Representations Improve AT}

\begin{wraptable}{r}{0.55\textwidth}
  \vspace{-0.5cm}
  \centering
  \caption{Clean and Robust Accuracy incorporating DRA on CIFAR-10, CIFAR-100, and ImageNet.}
  \label{tab_main_exp}
  \resizebox{\linewidth}{!}{%
  \begin{tabular}{ccccc}
    \toprule
    \multirow{2}{*}{Dataset} &
    \multirow{2}{*}{Model} &
    \multirow{2}{*}{Method} &
    \multirow{2}{*}{\makecell{Clean \\ Acc.}} &
    \multirow{2}{*}{\makecell{AutoAttack \\ Acc.}}\\
    \\
    \midrule
    \multirow{4.5}{*}{\makecell{CIFAR-10 \\ ($\ell_\infty=8/255$)}} &
      \multirow{2}{*}{WRN-28-10} & DM-AT       & 92.44 & 67.31   \\
     &                           & DM-AT + DRA & \textbf{93.17} & \textbf{67.86}  \\
    \cmidrule(lr){2-5}
     & \multirow{2}{*}{ViT-B/2}  & DM-AT       & 94.35 & 71.31  \\
     &                           & DM-AT + DRA & \textbf{95.13} & \textbf{72.03}  \\
         \midrule
    \multirow{4.5}{*}{\makecell{CIFAR-100 \\ ($\ell_\infty=8/255$)}} &
      \multirow{2}{*}{WRN-28-10} & DM-AT       & 68.34 & 35.72  \\
     &                           & DM-AT + DRA & \textbf{69.85} & \textbf{36.27}  \\
    \cmidrule(lr){2-5}
     & \multirow{2}{*}{ViT-B/2}  & DM-AT       & 68.53 & 36.52  \\
     &                           & DM-AT + DRA & \textbf{69.95} & \textbf{37.43}  \\
    \midrule
        \multirow{4.5}{*}{\makecell{ImageNet \\ ($\ell_\infty=4/255$)}} &
      \multirow{2}{*}{ConvNeXt-B} & DM-AT       & 74.40 & 54.34  \\
     &                           & DM-AT + DRA & \textbf{76.30} & \textbf{56.42}  \\
    \cmidrule(lr){2-5}
     & \multirow{2}{*}{ViT-B/16}  & DM-AT       & 74.67 & 54.88  \\
     &                           & DM-AT + DRA & \textbf{76.86} & \textbf{55.53}  \\
    \bottomrule
  \end{tabular}%
  }
  \vspace{-0.6cm}
\end{wraptable}

Table~\ref{tab_main_exp} summarizes the results on CIFAR-10, CIFAR-100, and ImageNet when combining DRA with the state-of-the-art DM-AT training recipe.
The results indicate that diffusion representations provide an effective feature prior for robust learning, leading to consistent gains in both clean accuracy and adversarial robustness.

We further evaluate CIFAR-10 across varying amounts of synthetic data (Table~\ref{tab_cifar10_only}), from none up to 50 million images.
While scaling synthetic data remains important, incorporating diffusion representations as an auxiliary learning signal more effectively leverages the robust knowledge encoded in diffusion models.
Moreover, compared to methods targeting the robust-overfitting regime (e.g., IKL, ADR),
which are reported to trade clean accuracy for robustness when training with diffusion synthetic data,
DRA improves DM-AT robustness without sacrificing clean accuracy.

Finally, we scale up to ImageNet (Table~\ref{tab_imagenet}) to verify that DRA remains effective when robust finetuning is applied to strong self-supervised foundation models on a downstream dataset,
with or without diffusion synthetic data.
Additional experiments with models trained from scratch are in Appendix~\ref{app:additional_in1k_exp}.

\begin{table*}[t]
  \caption{Comparison with state-of-the-art AT methods across different settings on CIFAR-10.}
  \vspace{-0.2cm}
  \label{tab_cifar10_only}
  \centering
  \scalebox{0.78}{
  \begin{tabular}{lcccllll}
    \toprule
     \multirow{2}{*}{\textbf{Method}} & \multirow{2}{*}{\textbf{Architecture}} & \multirow{2}{*}{\textbf{Syn.}} & \multirow{2}{*}{\textbf{\makecell{Epoch}}} & \multirow{2}{*}{\textbf{\makecell{Clean\\Acc.}}} & \multicolumn{3}{c}{\textbf{AutoAttack Acc.}} \\
     \cmidrule(lr){6-8}
     & & & & & $\ell_\infty\!=\!2/255$ & $\ell_\infty\!=\!8/255$ & $\ell_2\!=\!0.5$ \\
     \midrule
     AT & WRN-34-10 & - & 200 & 84.62 & 78.58 & 55.27 & 62.28 \\
    +ADR~\cite{wu2024annealing} & WRN-34-10 & - & 200 & 86.11 & 79.90 & 55.26 & 61.83 \\
     +IKL~\cite{cui2024dkl} & WRN-34-10 & - & 200 & 85.31 & 79.91 & 57.15 & 61.86 \\
     \rowcolor{gray!12} +DRA (Ours) & WRN-34-10 & - & 200 & $\mathbf{89.18_{\pm.11}}$ & $\mathbf{83.55_{\pm.10}}$ & $\mathbf{57.71_{\pm.20}}$ & $\mathbf{66.10_{\pm.44}}$\\
     \cmidrule(lr){1-8}
     DM-AT~\cite{wang2023better} & WRN-28-10 & 1M & 400 & 91.12 & 86.62 & 63.35 & 66.70 \\
     DM-AT~\cite{wang2023better} & WRN-28-10 & 1M & 800 & 91.43 & 87.17 & 63.96 & 66.72 \\
     \rowcolor{gray!12} +DRA (Ours) & WRN-28-10 & 1M & 400 & $\mathbf{92.32_{\pm.12}}$ & $\mathbf{87.90_{\pm.09}}$ & $\mathbf{64.09_{\pm.04}}$ & $\mathbf{69.19_{\pm.05}}$ \\
     \cmidrule(lr){1-8}
     DM-AT~\cite{wang2023better} & WRN-28-10 & 20M & 2400 & 92.44 & 88.38 & 67.31 & 70.25 \\
      +IKL~\cite{cui2024dkl} & WRN-28-10 & 20M & 2400 & 92.16 & 88.32 & 67.75 & 69.61 \\
      \rowcolor{gray!12} +DRA (Ours) & WRN-28-10 & 20M & 2400 & $\mathbf{93.17_{\pm.04}}$ & $\mathbf{89.42_{\pm.06}}$ & $\mathbf{67.86_{\pm.03}}$ & $\mathbf{71.77_{\pm.25}}$ \\
    \cmidrule(lr){1-8}
     DM-AT & ViT-B/2 & 20M & 500 & 92.21 & 88.21 & 66.41 & 70.08 \\
     \rowcolor{gray!12} +DRA (Ours) & ViT-B/2 & 20M & 500 & $\mathbf{93.26_{\pm.09}}$ & $\mathbf{89.54_{\pm.14}}$ & $\mathbf{67.68_{\pm.06}}$ & $\mathbf{72.52_{\pm.29}}$ \\
    \cmidrule(lr){1-8}
    DM-AT~\cite{wang2023better} & WRN-70-16 & 50M & 2000 & 93.25 & 89.71 & 70.69 & 69.78 \\
    +RA~\cite{peng2023robust} & RaWRN-70-16 & 50M & 2000 & 93.27 & 89.71 & 71.07 & 68.71 \\
    DM-AT~\cite{wu2025vision} & ViT-B/2 & 50M & 2000 & 94.35 & 91.10 & 71.31 & 72.00 \\
    \rowcolor{gray!12} +DRA (Ours) & ViT-B/2 & 50M & 2000 & $\mathbf{95.13_{\pm.06}}$ & $\mathbf{91.74_{\pm.02}}$ & $\mathbf{72.03_{\pm.18}}$ & $\mathbf{73.68_{\pm.14}}$ \\
    \bottomrule
  \end{tabular}
  }
  \vspace{-0.3cm}
\end{table*}

\begin{table}[t]
  \vspace{-0.2cm}
  \caption{Clean and Robust Accuracy incorporating DRA on ImageNet.}
  \centering
  \scalebox{0.8}{
  \begin{tabular}{cccccc}
    \toprule
    \multirow{2}{*}{\textbf{Method}} & \multirow{2}{*}{\textbf{Model}} & \multirow{2}{*}{\textbf{\makecell{Clean\\Acc.}}} & \multicolumn{3}{c}{\textbf{AutoAttack Acc.}} \\
    \cmidrule(lr){4-6}
    & & & $\ell_\infty\!=\!2/255$ & $\ell_\infty\!=\!4/255$ & $\ell_2\!=\!2.0$ \\
    \midrule

    \multirow{6}{*}{\shortstack{AT \\ ($\ell_\infty\!=\!4/255$)}}    & ConvNeXt-B & $71.91_{\pm.48}$ & $61.31_{\pm.54}$ & $49.40_{\pm.44}$ & $40.01_{\pm.81}$ \\
                           & +DRA & $\mathbf{74.53_{\pm.10}}$ & $\mathbf{64.17_{\pm .07}}$ & $\mathbf{51.77_{\pm .12}}$ & $\mathbf{44.65_{\pm .76}}$ \\
                           & \cellcolor{gray!12}\textcolor{darkgray}{\small $\Delta$} & \cellcolor{gray!12}\textcolor{darkgreen}{\small $+2.62$} & \cellcolor{gray!12}\textcolor{darkgreen}{\small $+2.86$} & \cellcolor{gray!12}\textcolor{darkgreen}{\small $+2.37$} & \cellcolor{gray!12}\textcolor{darkgreen}{\small $+4.64$} \\
                           \cmidrule(lr){2-6}
                           & ViT-B/16   & $73.75_{\pm.23}$ & $63.22_{\pm.09}$ & $51.62_{\pm.02}$ & $53.26_{\pm.23}$ \\
                           & +DRA & $\mathbf{76.23_{\pm.04}}$ & $\mathbf{65.64_{\pm.01}}$ & $\mathbf{52.95_{\pm.03}}$ & $\mathbf{55.41_{\pm.05}}$ \\
                           & \cellcolor{gray!12}\textcolor{darkgray}{\small $\Delta$} & \cellcolor{gray!12}\textcolor{darkgreen}{\small $+2.48$} & \cellcolor{gray!12}\textcolor{darkgreen}{\small $+2.42$} & \cellcolor{gray!12}\textcolor{darkgreen}{\small $+1.33$} & \cellcolor{gray!12}\textcolor{darkgreen}{\small $+2.15$} \\
    \cmidrule(lr){1-6}
    \multirow{6}{*}{\shortstack{DM-AT \\ ($\ell_\infty\!=\!4/255$)}} & ConvNeXt-B & $74.40_{\pm.09}$ & $65.26_{\pm.26}$ & $54.34_{\pm.09}$ & $43.54_{\pm.18}$ \\
                           & +DRA & $\mathbf{76.30_{\pm.18}}$ & $\mathbf{67.48_{\pm.16}}$ & $\mathbf{56.42_{\pm.23}}$ & $\mathbf{47.64_{\pm.23}}$ \\
                           & \cellcolor{gray!12}\textcolor{darkgray}{\small $\Delta$} & \cellcolor{gray!12}\textcolor{darkgreen}{\small $+1.90$} & \cellcolor{gray!12}\textcolor{darkgreen}{\small $+2.22$} & \cellcolor{gray!12}\textcolor{darkgreen}{\small $+2.08$} & \cellcolor{gray!12}\textcolor{darkgreen}{\small $+4.10$} \\
                           \cmidrule(lr){2-6}
                           & ViT-B/16   & $74.67_{\pm.05}$ & $65.65_{\pm.05}$ & $54.88_{\pm.16}$ & $50.75_{\pm1.03}$ \\
                           & +DRA & $\mathbf{76.86_{\pm.06}}$ & $\mathbf{67.44_{\pm.12}}$ & $\mathbf{55.53_{\pm.22}}$ & $\mathbf{55.81_{\pm.92}}$ \\
                           & \cellcolor{gray!12}\textcolor{darkgray}{\small $\Delta$} & \cellcolor{gray!12}\textcolor{darkgreen}{\small $+2.19$} & \cellcolor{gray!12}\cellcolor{gray!12}\textcolor{darkgreen}{\small $+1.79$} & \cellcolor{gray!12}\textcolor{darkgreen}{\small $+0.65$} & \cellcolor{gray!12}\textcolor{darkgreen}{\small $+5.06$} \\
    \bottomrule
  \end{tabular}
  }
  \label{tab_imagenet}
  \vspace{-0.5cm}
\end{table}

\subsection{Diffusion Representation Alignment Improves Representation Quality}

To understand if DRA actually improves representation quality, we plot the uniformity and alignment metrics on CIFAR-10 trained checkpoints, along with the corresponding clean and robust accuracy.
Figure~\ref{fig:uni_align_scatter} presents the results, showing that DRA effectively leverages the diverse features encoded in diffusion representations, contributing to the improved clean and robust accuracy.

\subsection{Diffusion Models Encourage Learning of More Easily Disentangled Representations}

Recent mechanistic interpretability work hypothesizes that models may rely on feature superposition to encode more features than the available representation dimensions,
which involves representing features as non-orthogonal directions in the activation space~\cite{elhage2022superposition}.
While this can be effective for compressing features that rarely co-activate, it has been suggested that superposition may be exploited by adversarial examples~\cite{gorton2025advsuper, stevinson2025adversarial}.

Moving from toy settings to real-world datasets, where the degree of superposition is difficult to quantify, \citet{gorton2025advsuper} use sparse autoencoders (SAEs)~\cite{huben2024sparse} as a proxy for understanding the effect of robust training on classifier representations.
SAEs learn a set of wide but sparse and interpretable features that aim to reconstruct model activations, with lower reconstruction loss reflecting that the representation is easier to disentangle into the set of sparse features.
To investigate the effect of incorporating diffusion models into adversarial training, we train TopK-SAEs on classifier representations with different sparsity levels, $K\in \{8, 16, 32\}$, using model checkpoints trained on ImageNet, and compare the normalized SAE reconstruction loss~\cite{gao2025scaling, gorton2025advsuper}.

\begin{figure}[t]
    \centering
    \begin{minipage}[t]{0.48\textwidth}
        \centering
        \includegraphics[width=\linewidth]{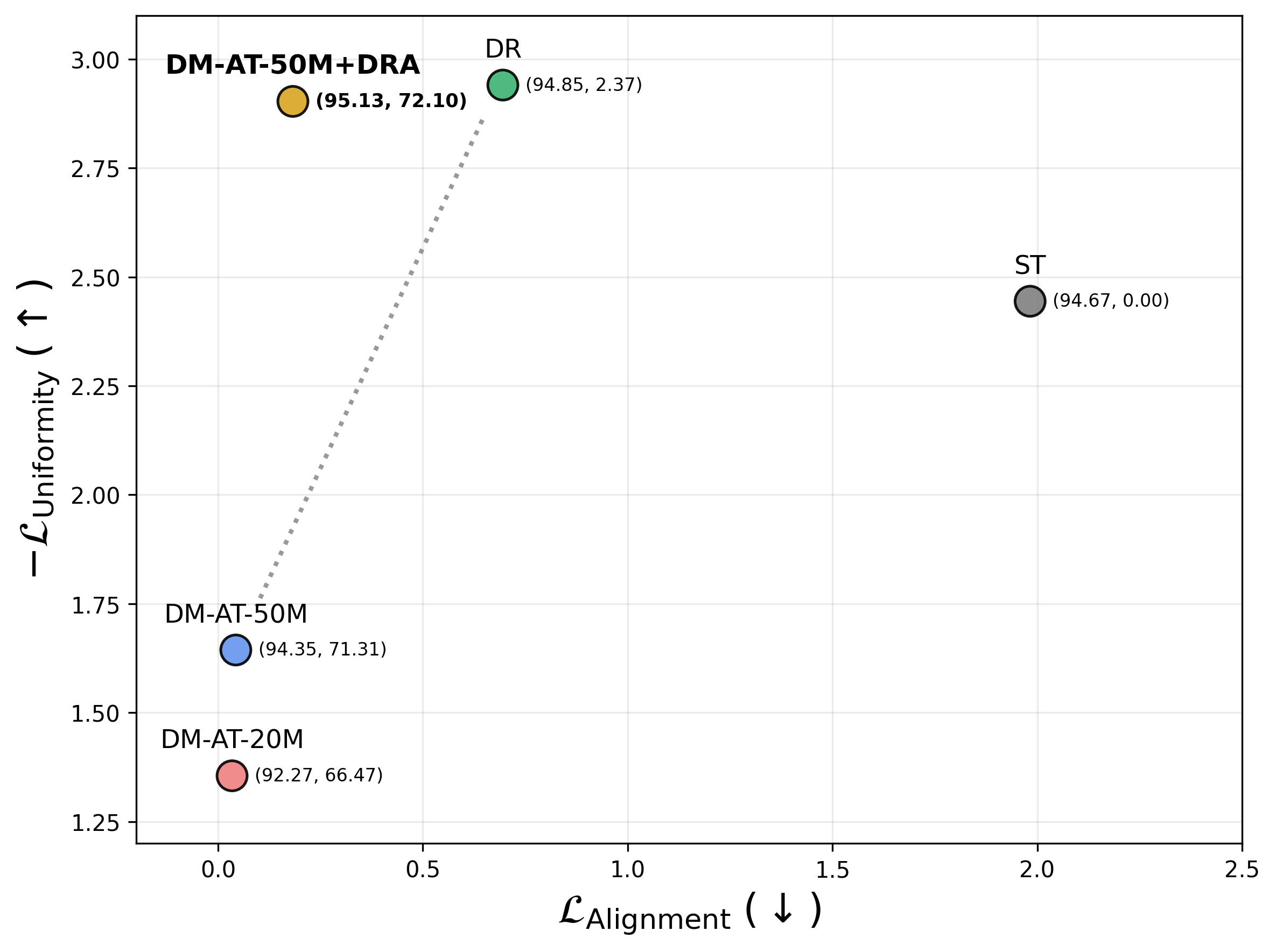}
        \vspace{-0.3cm}
        \captionof{figure}{We plot the alignment and uniformity metrics, along with clean and robust accuracy on CIFAR-10 ($\ell_\infty=8/255$) shown in parentheses.}
        \label{fig:uni_align_scatter}
    \end{minipage}
    \hfill
    \begin{minipage}[t]{0.48\textwidth}
        \centering
        \includegraphics[width=\linewidth]{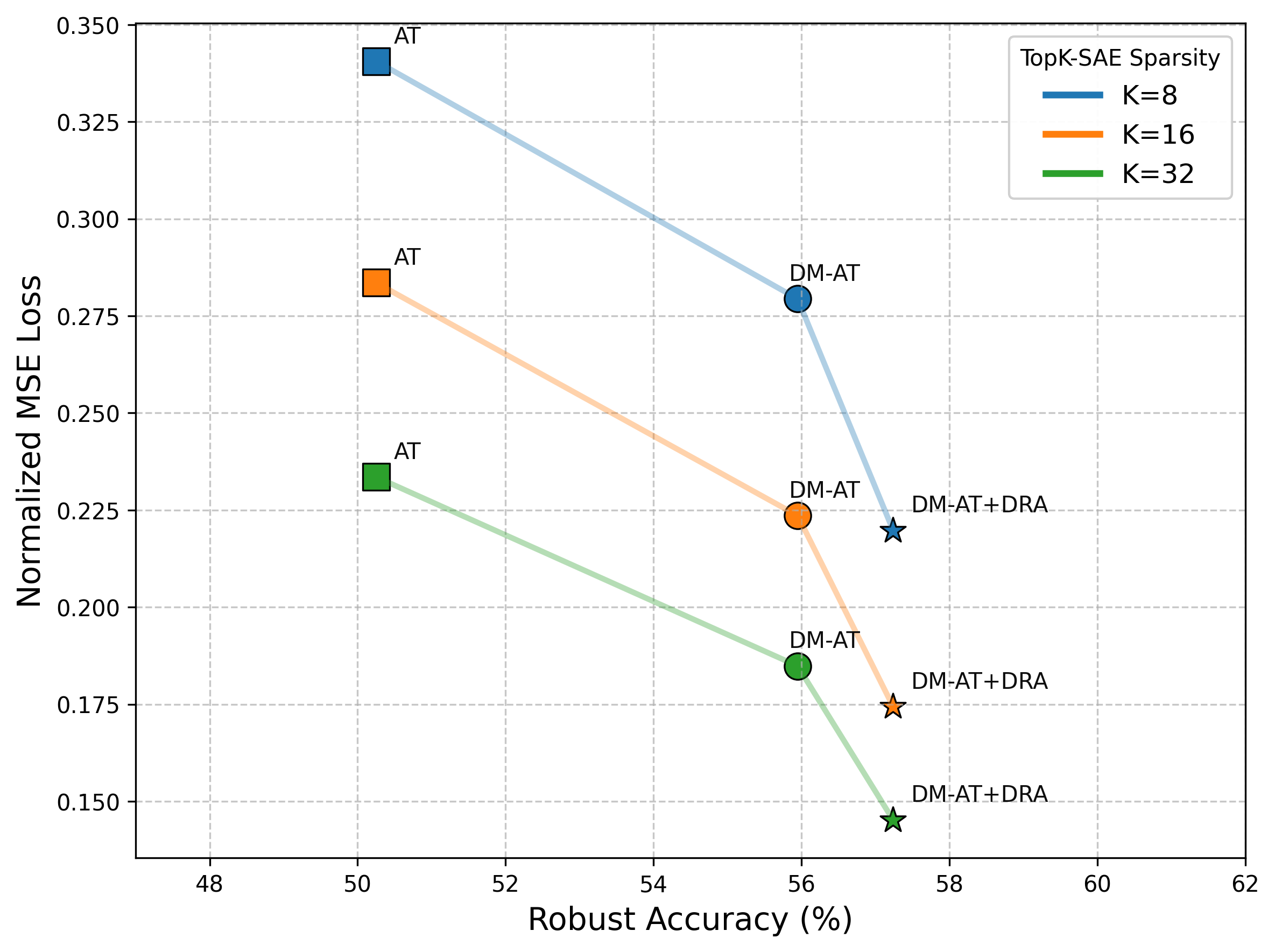}
        \vspace{-0.3cm}
        \captionof{figure}{Normalized reconstruction losses of TopK SAEs trained on ImageNet using AT, DM-AT, and DM-AT+DRA ViT-B checkpoints.}
        \label{fig:sae_loss}
    \end{minipage}
    \vspace{-0.4cm}
\end{figure}

Figure~\ref{fig:sae_loss} presents the results.
We find that incorporating diffusion synthetic data and diffusion representation alignment improves robustness while also reducing the reconstruction loss, suggesting that the learned representations become easier to decompose into sparse features.
This complements the observations of \citet{gorton2025advsuper}, who report that adversarial training encourages more easily disentangled representations, with larger perturbation budgets further amplifying this effect.
Our results show that incorporating diffusion models into adversarial training can improve robustness and also encourage models to learn representations that are easier to disentangle.

\subsection{Distinct Roles of Diffusion Models}
\label{sec:distinct_roles}

Diffusion synthetic data augments the pool of input examples for adversarial training, while diffusion representation alignment provides an auxiliary learning signal from the mildly robust and diverse feature prior.
In this section, we further investigate how incorporating these methods into robust training affects the resulting robust classifiers.

To understand how they shape classifier representations, we build on the methodology of~\citet{feng2022rank} and examine whether the use of representation dimensions changes when diffusion synthetic data and diffusion representations are introduced. Specifically,~\citet{feng2022rank} propose \emph{classification dimension} as a measure of the intrinsic dimensionality of the feature space, approximated by the minimum number of principal components needed to preserve high classification accuracy.

Concretely, we first apply PCA to the classifier representations to obtain eigenvectors. We then modify the forward pass by projecting the representation onto the top-$K$ eigenvectors before feeding it into the classification head, and measure the resulting accuracy.
As a robust-aware variant, we additionally measure robust accuracy by computing eigenvectors from clean representations, and projecting adversarial representations onto the subspace.

Figure~\ref{fig:cls_dim_example} (Appendix~\ref{app:cls_dim_example}) shows an example on a robust model trained on CIFAR-10. As expected, classification accuracy gradually recovers to the original performance as more eigenvectors are included. For robust accuracy, however, we observe that performance first improves and then degrades as $K$ increases, suggesting that adversarial perturbations may disproportionately exploit less important principal components.
Furthermore, Table~\ref{tab:cls_dim_results} presents the results for CIFAR-10 adversarially trained models with the minimum number of principal components required to recover 99\% of the original clean accuracy (CLS Dim), and the robust-aware dimension that achieves the highest robust accuracy (Robust Dim).
The results suggest that synthetic data concentrates predictive information in fewer principal components, while DRA increases the number needed to preserve clean accuracy.

\begin{wraptable}{r}{0.48\textwidth}
\vspace{-0.5cm}
\centering
\caption{Classification and robust dimensions with WRN-28-10s on CIFAR-10.}
\label{tab:cls_dim_results}
\scalebox{0.85}{
\begin{tabular}{lcc}
\toprule
\textbf{Method} & \textbf{CLS Dim} & \textbf{Robust Dim} \\
\midrule
AT & 14 & 9 \\
AT+DRA & 42 & 22 \\
\cmidrule(lr){1-3}
DM-AT  & 11 & 11 \\
DM-AT+DRA  & 15 & 23 \\
\bottomrule
\end{tabular}
}
\vspace{-0.4cm}
\end{wraptable}

While it is intuitive that representation alignment may promote diverse and robust feature encoding,
the reduction in the number of components needed for classification when trained with synthetic data may be related to the observation that diffusion synthetic examples are easier to classify than the original real data~\cite{hu2024a}.
Prior work has also explored using partially synthesized diffusion data as an augmentation strategy to improve robustness~\cite{sastry2024diffaug}, though not specifically in the context of enhancing adversarial training recipes.
Although previous work has largely attributed the benefits of diffusion-based synthetic data to improved image quality \cite{wang2023better, bartoldson2024adversarial}, it might not be the most essential factor behind its effectiveness.
We leave this viewpoint as a potential direction for further improving adversarial training.

\subsection{Ablation Studies}
\label{exp:ablation}

\begin{wrapfigure}{r}{0.42\textwidth}
    \vspace{-1.5cm}
    \centering
    \includegraphics[width=\linewidth]{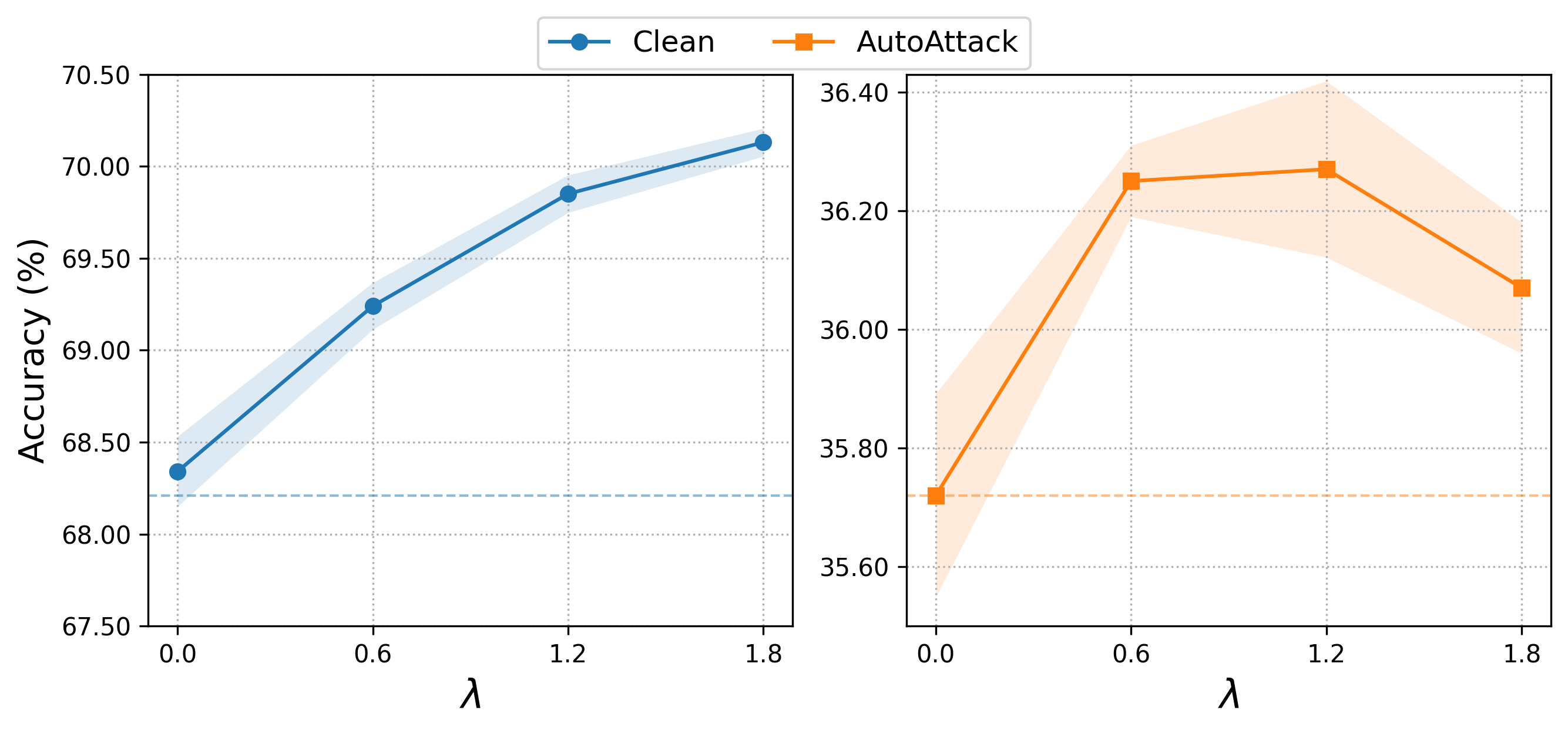}
    \caption{Clean and Robust accuracy on CIFAR-100 for DM-AT+DRA with different DRA regularization strength $\lambda$.}
    \label{fig:reg_strength}
    \vspace{-0.5cm}
\end{wrapfigure}
\textbf{Regularization Strength.}

In our experiments, the alignment regularization strength is set to $\lambda=1.2$.
We select this value based on an initial sweep using WRN-28-10 on CIFAR-100, and then fix the same coefficient for the remaining settings to reduce computational cost.
Figure~\ref{fig:reg_strength} shows that DRA consistently improves upon the baseline in both clean and robust accuracy.
When $\lambda$ becomes too large, robustness improvements saturate, as the alignment term can outweigh the original adversarial training objective.

\paragraph{Does Noisy-Input Training Alone Learn Good Enough Features?}

Given that representations extracted from diffusion models serve as effective feature priors for robust learning, a natural question is whether standard training coupled with the same noisy-input procedure is sufficient to achieve similar benefits.
We explore this hypothesis by applying noisy-input discriminative pre-training to the same UNet encoder architecture used in the EDM diffusion model, using this encoder as the alignment target in Figure~\ref{fig:method_overview}.
Results show that replacing the diffusion target in a WRN-28-10 trained with DM-AT+DRA (1M synthetic) reduces robust accuracy from 64.09\% to 62.62\%, which is inferior to the vanilla DM-AT baseline.
This indicates that noisy-input training alone is insufficient; rather, the generative training objective of diffusion models is a critical factor in producing feature priors that benefit downstream robust training.
Relatedly, \citet{jaini2024intriguing} find that diffusion-like noisy discriminative pre-training increases shape bias but hurts OOD accuracy.
As a complementary finding, we analyze the frequency behavior of leveraging diffusion representations or the noisy-input pretrained discriminative features, and find that the latter have an undesirable effect of increasing the sensitivity to mid-high frequency components (Appendix~\ref{app:noisy_disc_exp}).

\paragraph{Why Not Adversarially Finetune the Diffusion Encoder Directly?}
Several works in self-supervised learning that leverage diffusion representations finetune the diffusion encoder with a prediction head on top~\cite{xiang2023ddae, yagoda2025diffrobust}.
We evaluate this approach in our initial exploration under the same AT setting as in Table~\ref{tab_cifar10_only}. It achieves 87.77\% clean accuracy and 55.76\% robust accuracy.
While this improves over the WRN-34-10 AT baseline, it is still below WRN-34-10 AT+DRA and is less training-efficient (1.35$\times$ training time per epoch).
Additionally, diffusion models are often trained with class conditioning for data generation. However, deploying the diffusion encoder at inference time can only provide unconditional representations, which can result in a slight decrease in representation quality depending on the available diffusion model~\cite{xiang2023ddae, chen2025deconstructing}.
In contrast, DRA better leverages the robust knowledge encoded in diffusion representations while enabling more flexible downstream classifier choices for the classification task.

Further ablations and analyses, including DRA computational cost, alternative SSL feature alignment targets, and SAE-based representation analysis, are provided in the Appendix.

\section{Conclusions}

We systematically explore whether diffusion models can improve adversarial training beyond synthetic data generation alone.
We find that diffusion representations are partially robust yet diverse, and show that aligning classifiers with them as an auxiliary learning signal consistently improves AT across settings and datasets.
Our analysis further shows that incorporating diffusion models encourages representations that are easier to disentangle.
We hope these findings inspire further uses of diffusion models in AT beyond generating high-quality synthetic images.

\newpage

\section*{Acknowledgement}

We thank the anonymous reviewers and members of CLLab and AISLab for their constructive feedback. This work was supported in part by the National Science and Technology Council in Taiwan under Grants NSTC 113-2222-E-002-004-MY3, 114-2634-F-002-004, 115-2628-E-002-029, 114-2221-E-002-102-MY3, the Ministry of Education under Grant 114M7091, NTU AI Center of Research Excellence within Taiwan Centers of Excellence, and NTU Center for Data Intelligence. We thank the National Center for High-performance Computing (NCHC) of National Institutes of Applied Research (NIAR) in Taiwan for providing computational and storage resources. H.-T. Lin is honored to be supported by the Leap Fellowship of the Foundation for the Advancement of Outstanding Scholarship in Taiwan since 2025.

\bibliographystyle{plainnat}
\bibliography{mybib}

\clearpage
\etocsettagdepth{mtchapter}{none}
\etocsettagdepth{appendix}{subsection}
\etocsetnexttocdepth{subsection}
\newpage
\appendix
\clearpage

\etocruledstyle[1]{\Large\bfseries Appendix}
{\hypersetup{linkcolor=black}\tableofcontents}
\clearpage

\etocdepthtag.toc{appendix}

\section{Representation Similarity Analysis}
\label{app:imp_alignment_trend}
\begin{figure}[ht]
    \centering
    \includegraphics[width=0.5\textwidth]{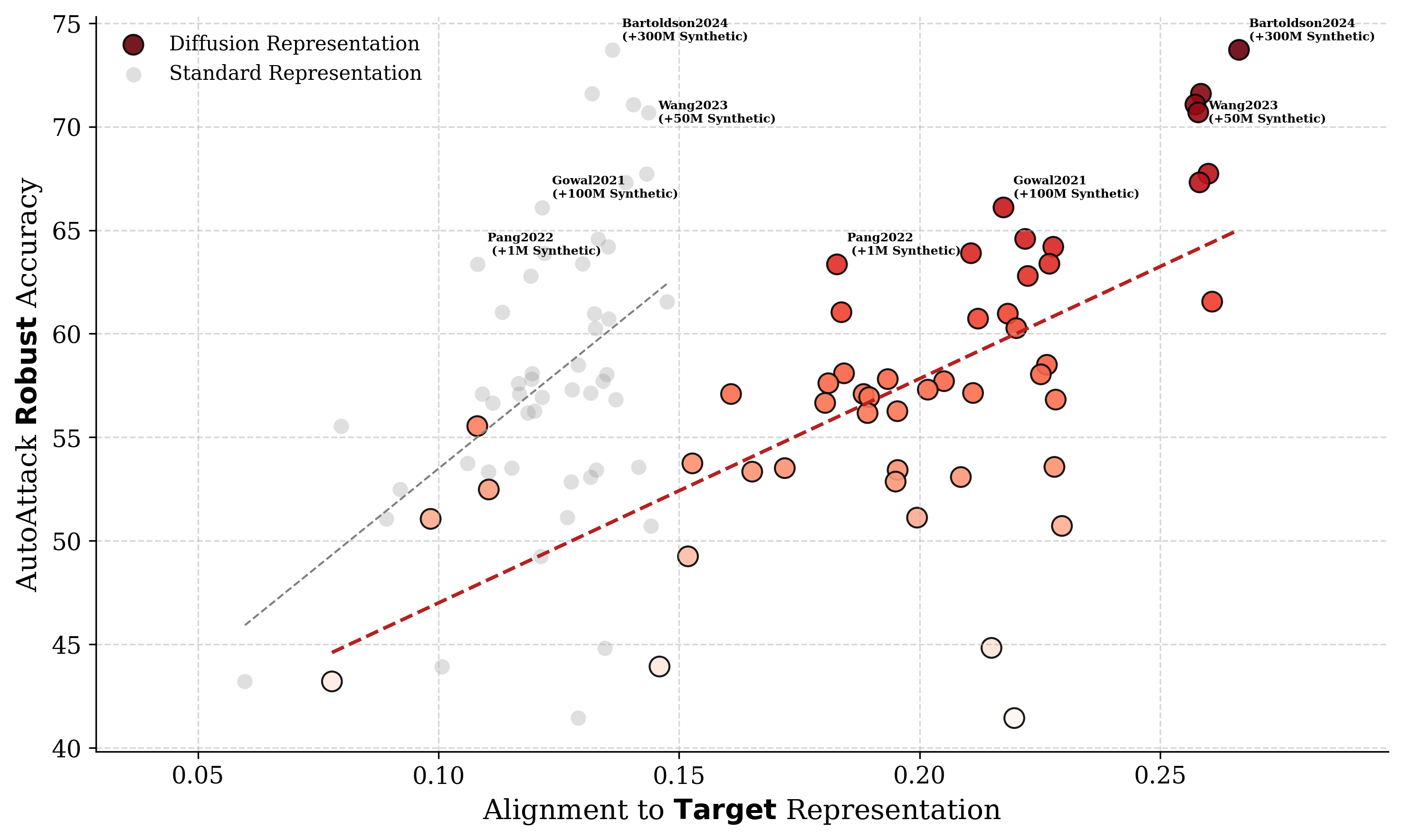}
    \caption{ 
    Similarity scores are measured with respect to representations extracted from the diffusion model (\textcolor{darkred}{red}) and from the RobustBench standard-trained model (\textcolor{darkgray}{gray}).
    }
    \label{fig:alignment_trend_full}
    \vspace{-0.15cm}
\end{figure}

As a motivational experiment, following the work of~\citet{huh2024position} on analyzing representation similarities between models trained with different learning objectives, we extract the representations from the CIFAR-10 EDM diffusion model~\cite{karras2022edm} used in DM-AT~\cite{wang2023better}, and measure the representation similarity score CKNNA~\cite{huh2024position} with the CIFAR-10 $\ell_\infty$-robust models from RobustBench~\cite{croce2021robustbench}. 
Figure~\ref{fig:align_fig_1} presents the results, showing that better robust models, often trained with abundant synthetic images, already have the tendency to exhibit higher representation similarity with diffusion representations.
As the tendency could also be explained by better robust models having improved natural classification performance for clean images, we also plot the representation similarity score with the standard-trained model from RobustBench (Figure~\ref{fig:alignment_trend_full}).
The observed correlation motivates us to investigate whether explicitly leveraging diffusion representations as an auxiliary learning signal is beneficial for robust classifier training.

\section{Motivational Frequency Analysis Implementation and Additional Results}

\textbf{Implementation.} We perform the frequency analysis on $\ell_\infty$ PGD-20 input
perturbations ($\varepsilon = 8/255$ on CIFAR-10 and $4/255$ on ImageNet).
Following prior frequency-saliency analyses~\cite{huang2023improving,chan2022how}, we compute
the magnitude of the 2-D discrete Fourier transform of each perturbation, average
it over images and channels, shift the low frequencies to the center, and normalize
each spectrum to $[0,1]$ so that models can be compared.

\begin{figure}[ht]
    \centering
    \includegraphics[width=0.7\textwidth]{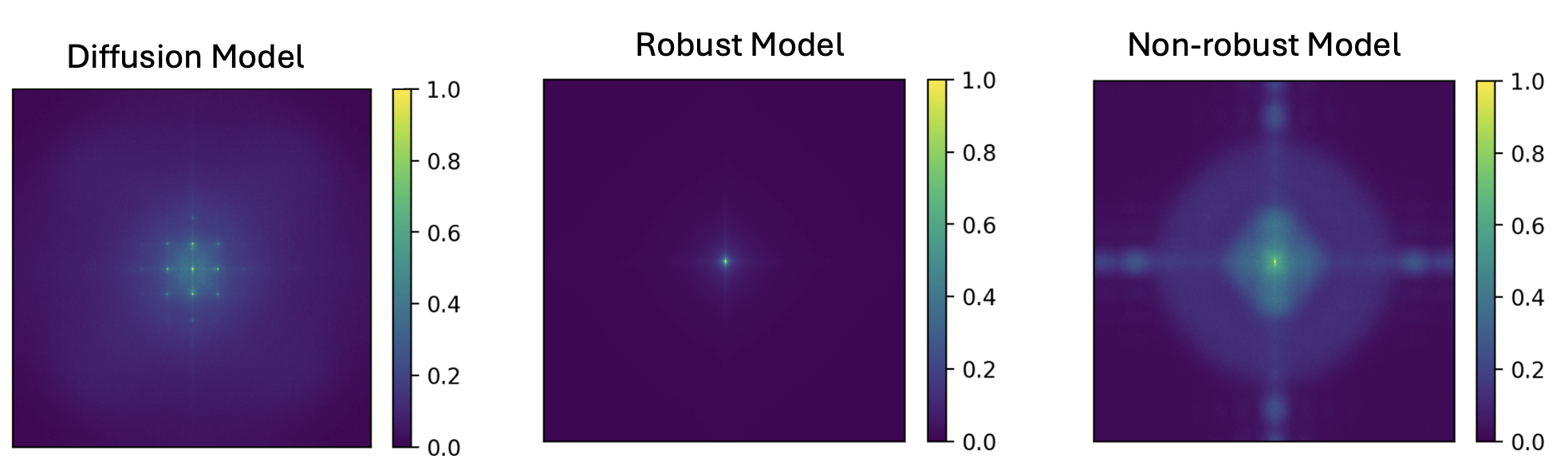}
    \caption{ 
        The frequency saliency analysis of the linearly probed diffusion representation, adversarially trained robust model, and standard non-robust model. Low frequencies are centered.
    }
    \label{fig:freq_map_in1k}
    \vspace{-0.15cm}
\end{figure}

In addition to Figure~\ref{fig:freq_map}, Figure~\ref{fig:freq_map_in1k} reports
the same analysis on ImageNet for a linear probe on frozen LightningDiT
representations, a standard-trained non-robust ViT-B/16, and an
adversarially trained robust ViT-B/16.
The standard-trained model exhibits
less high-frequency reliance than its CIFAR-10 counterpart, but it still shows the
strongest mid-frequency reliance of the three.
Note that the diffusion model in this setting is a latent diffusion model, so the perturbation is also affected by the VA-VAE encoder, which may contribute to the frequency reliance in this setting.

\section{Experiment Setup Details}

\label{app:exp_setup_details}
\textbf{Training Setup.} For CIFAR-10/100 WideResNet training, we follow the exact same setup as DM-AT~\cite{wang2023better}: PGD-10 TRADES~\cite{zhang2019trades} with $\beta=5$, weight averaging with $\tau=0.995$, SGD with momentum 0.9, weight decay $5\times10^{-4}$, and $lr=0.2$ with a cosine annealing scheduler.
For ViT-B/2, we follow~\citet{wu2025vision} and use the Lion optimizer~\cite{chen2023lion} with batch size 1024, $lr=10^{-4}$, and weight decay 0.5.
For ImageNet models, we fully finetune for 100 epochs using the Lion optimizer and the common PGD-3 TRADES with $\beta=10$.
For the baselines, we report the numbers from the original paper or evaluate them with the released checkpoints.
For our experiments, statistics over three runs are reported.

For the SAE experiments, we follow prior works~\cite{gorton2025advsuper, gao2025scaling} using TopK-SAEs trained with an expansion factor of $8$, $lr=5\times10^{-4}$, and $k_{aux}=512$ with auxiliary weight $0.1$, using the Adam optimizer~\cite{kingma2015adam} for a total of 50000 steps.

\textbf{Computational Resources.}
The experiments are mostly conducted on machines with 4 GPUs (NVIDIA H100 80GB), with 12 CPU cores and 200GB of RAM per GPU.
Each training run using 4 GPUs takes around 2 minutes per epoch for the CIFAR experiments, and 20 minutes per epoch for the ImageNet experiments.
Due to the expensive nature of adversarial training and evaluations, an estimated total of 6000 GPU hours is needed for the experiments, including training runs, evaluations, and preliminary explorations.

\section{Additional Discussion for Related Work}
\label{app:det_for_related_work}

\textbf{Frozen Diffusion Features for Robustness.} \citet{yagoda2025diffrobust} show that training a classification head on frozen, unconditional diffusion encoders can achieve robustness that is slightly below adversarially trained models, while being much cheaper to train.
However, as their approach relies on inference-time randomness, robust evaluation should be more carefully considered~\cite{athalye2018obfuscated}.
For example, the configuration of Attention Head with hyperparameters $b$=8 and $t$=50 has been reported to achieve 46.0\% AutoAttack robust accuracy, but a simple EOT-based evaluation could reduce the reported robust accuracy to 17.3\%.
We also evaluate a linear probe on an EDM diffusion model on CIFAR-10 by fixing the added noise to remove inference-time randomness during adversarial evaluation (Figure~\ref{fig:robust_acc}). The results show that diffusion representations still require robust training to provide competitive robustness.

\textbf{Representation Alignment with Diffusion Models.}
A growing line of work inspects diffusion features for a variety of downstream tasks via feature alignment.
CleanDIFT~\citep{stracke2025cleandift} leverages self-distillation across timesteps to produce clean, timestep-independent features for dense prediction tasks;
DreamTeacher~\citep{li2023dream} and RepFusion~\citep{yang2023diffusion} distill diffusion priors into discriminative visual models for downstream discriminative tasks;
REPA~\citep{yu2025repa} aligns diffusion intermediate features with self-supervised features to accelerate the training of diffusion transformers.
These works converge on a similar recipe: a projection head with a distillation loss such as cosine similarity or L2 distance to align the feature space.
In this work, we build upon the alignment mechanism, but instead study whether diffusion representations are beneficial as an auxiliary learning signal for robust classifier training in AT.

\section{Representation Alignment Implementation Details}
\label{app:DRA_Imp_Abl}

\textbf{Diffusion Representation Extraction.}
For CIFAR-10/100, we use the same EDM diffusion model checkpoint as \citet{wang2023better} to generate synthetic images.
We extract representations at noise scale $\sigma_t=0.1$ from the UNet bottleneck block before the upsampling layers, and apply average pooling over spatial dimensions.

For ImageNet, we use the LightningDiT checkpoint released by \citet{yao2025ldit}. We extract features from middle layer 14 at $t=0.8$, where $t=0$ corresponds to pure noise and $t=1$ to the clean image.
We also found that LightningDiT representations suffer from the large-activation issue reported in prior work \cite{fang2025tiny, gan2025unleashing}, which reduces their effectiveness as a learning signal.
To mitigate this, we apply the method introduced in \citet{gan2025unleashing},
which applies the zero-out-and-AdaLN-normalize method on the channels with abnormally large activations (channels 1053 and 259 in the released LightningDiT checkpoint).
Finally, average pooling over spatial dimensions is applied to extract the representations.

Lastly, for CIFAR-10 models trained with more than one million synthetic images, we extract an additional representation per image by sampling an extra timestep using the same sampling function the EDM model originally used during training, and take the average of the cosine similarity loss.
This results in a slight improvement, likely by better leveraging the representations across timesteps~\cite{stracke2025cleandift, li2025understanding}.

\textbf{Alignment Module.} We follow \citet{yu2025repa} and use a 3-layer MLP, trained with the cosine similarity loss.
In initial experiments, more sophisticated alignment heads and loss functions did not improve performance, so we retain this simple configuration as recommended in prior work \cite{yu2025repa, stracke2025cleandift}.

\section{Computational Cost of DRA}
\label{app:compute_cost}
Since the computational cost is dominated by the multiple forward and backward passes required for adversarial training, extracting features from the diffusion encoder on the fly at each iteration introduces minimal training-time overhead ($\sim\!1.04\times$ training time with $\sim\!1$ GB additional VRAM per GPU on CIFAR-10/100, and $\sim\!3$ GB additional VRAM with $\sim\!1.13\times$ training time per GPU for our ImageNet experiments).
In practice, one could consider further reducing the training time and memory overhead by precomputing the diffusion features during the generation of the synthetic data for DM-AT, albeit at the cost of increased disk storage.

\section{Additional Analysis on SAE Features}
\label{app:sae_analysis}

\begin{figure}[ht]
    \centering
    \begin{subfigure}[t]{0.48\linewidth}
        \centering
        \includegraphics[width=\linewidth]{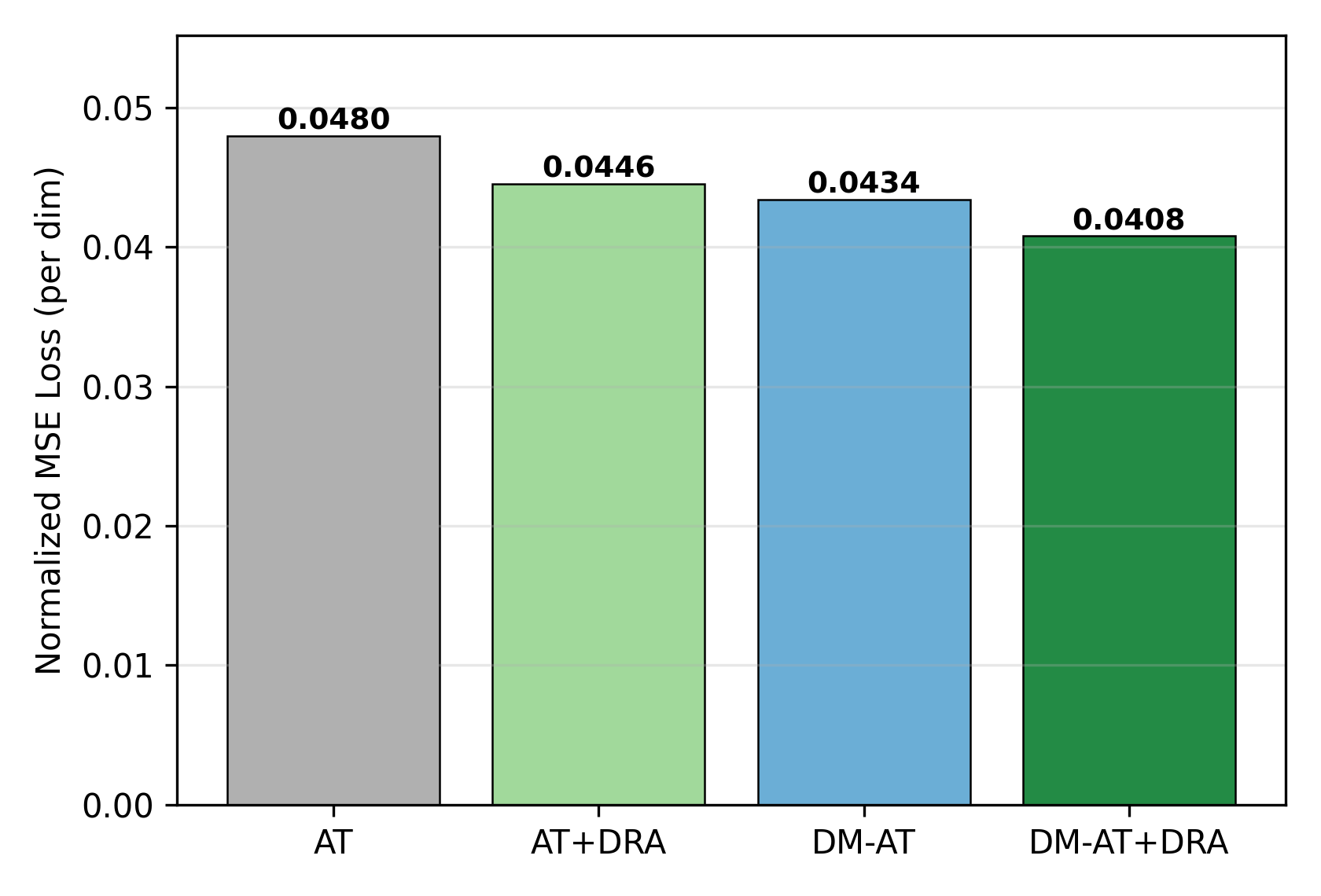}
        \caption{Normalized TopK-SAE reconstruction loss.}
        \label{fig:cifar10_sae_loss}
    \end{subfigure}\hfill
    \begin{subfigure}[t]{0.48\linewidth}
        \centering
        \includegraphics[width=\linewidth]{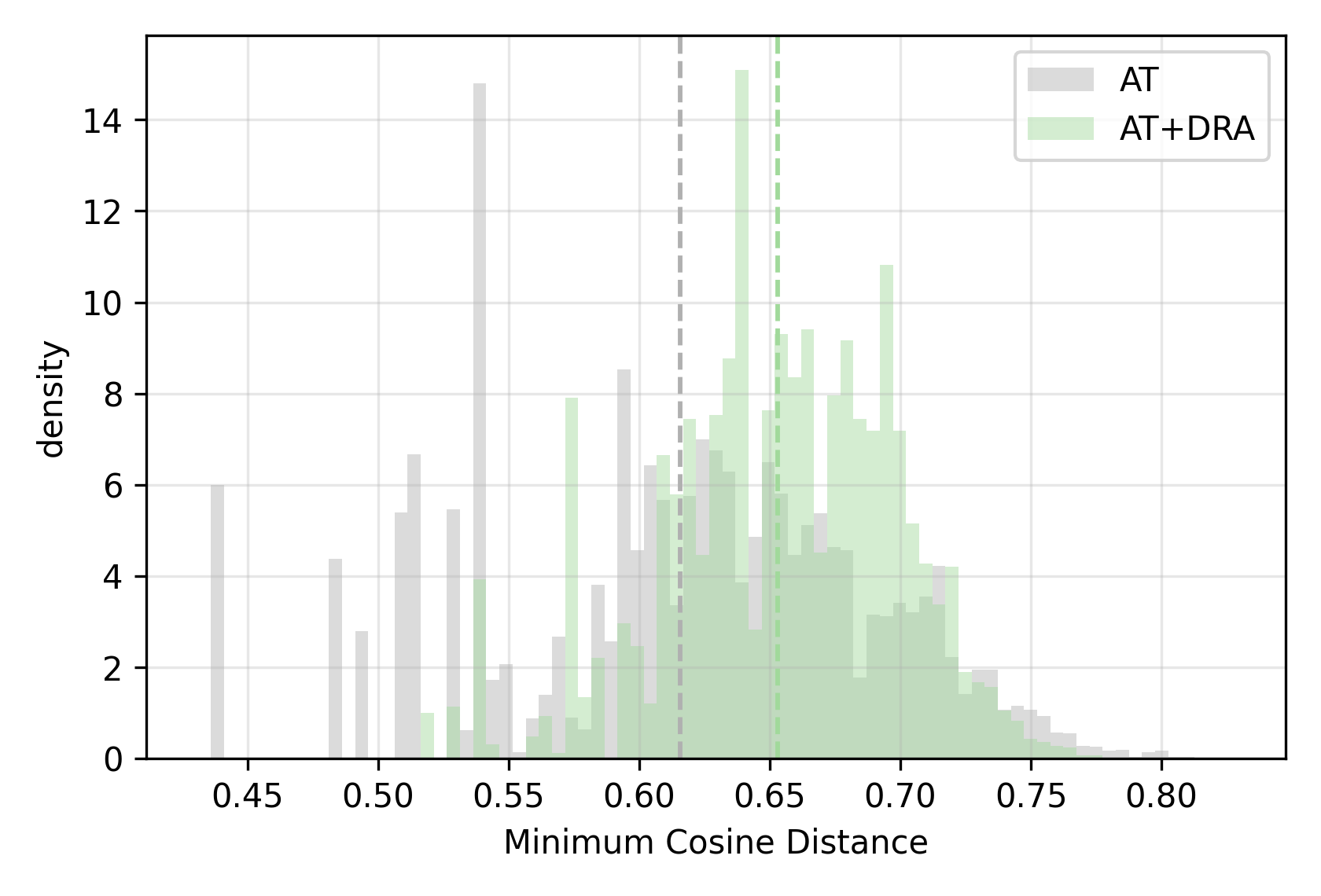}
        \caption{AT vs.~AT+DRA.}
        \label{fig:sae_cosine_at_atdra}
    \end{subfigure}

    \vspace{0.3cm}

    \begin{subfigure}[t]{0.48\linewidth}
        \centering
        \includegraphics[width=\linewidth]{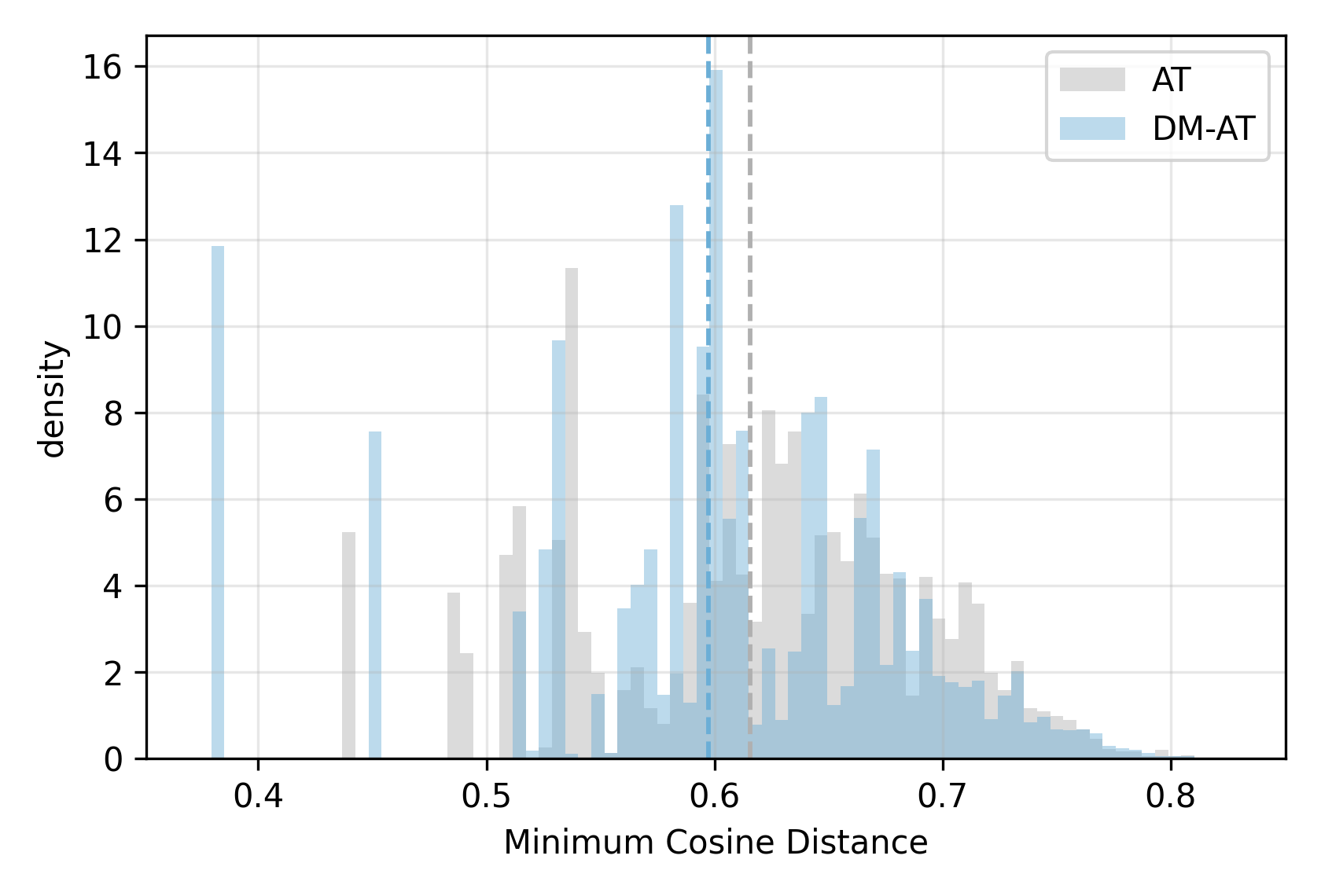}
        \caption{AT vs.~DM-AT.}
        \label{fig:sae_cosine_at_dmat}
    \end{subfigure}\hfill
    \begin{subfigure}[t]{0.48\linewidth}
        \centering
        \includegraphics[width=\linewidth]{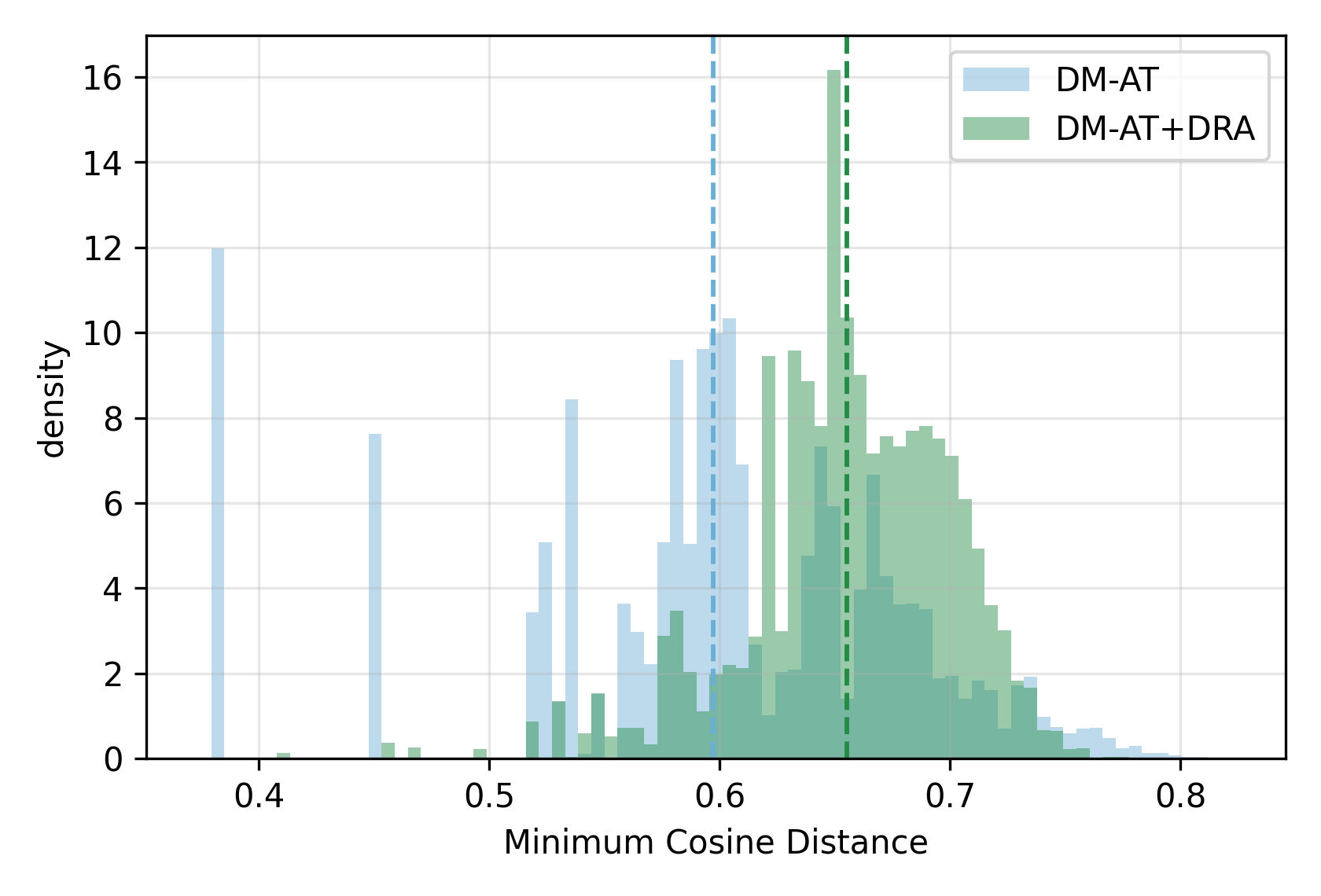}
        \caption{DM-AT vs.~DM-AT+DRA.}
        \label{fig:sae_cosine_dmat_dmatdra}
    \end{subfigure}
    \caption{SAE analyses on the same CIFAR-10 checkpoints used in Section~\ref{sec:distinct_roles}. (a) Normalized TopK-SAE reconstruction loss across AT, AT+DRA, DM-AT, and DM-AT+DRA. (b)--(d) Distributions of the minimum pairwise cosine distance among the TopK (K=32) active SAE features per input. Dashed lines mark distribution means.}
    \label{fig:sae_cifar10}
\end{figure}

Using the same CIFAR-10 checkpoints analyzed in Section~\ref{sec:distinct_roles}, we further examine the SAE features that the classifier uses to reconstruct each input's representation.

\textbf{Reconstruction loss on CIFAR-10.}
Figure~\ref{fig:cifar10_sae_loss} presents the normalized TopK-SAE ($K\!=\!32$) reconstruction loss on CIFAR-10 checkpoints. 
Similar patterns observed on ImageNet hold, where incorporating diffusion models encourages more disentangled representations.

\textbf{Directional diversity of active features.}
We additionally probe the directional structure of the active features. For each input, we take the TopK (K=32) active features used to reconstruct its representation and compute the minimum pairwise cosine distance among them.
A larger minimum distance indicates that the model recruits a more directionally diverse set of features to encode the representation of that input.

Figures~\ref{fig:sae_cosine_at_atdra}--\ref{fig:sae_cosine_dmat_dmatdra} report the resulting distributions.
Adding DRA shifts the distribution toward larger minimum cosine distances, both on top of AT and on top of DM-AT.
In contrast, switching from AT to DM-AT does not produce a comparable shift.
This complements the analysis in Section~\ref{sec:distinct_roles}, showing that using DRA encourages the per-input active SAE features to point in more distinct directions.

\section{Classification Dimension Visualization}
\label{app:cls_dim_example}
\begin{figure}[ht]
    \centering
    \includegraphics[width=0.45\textwidth]{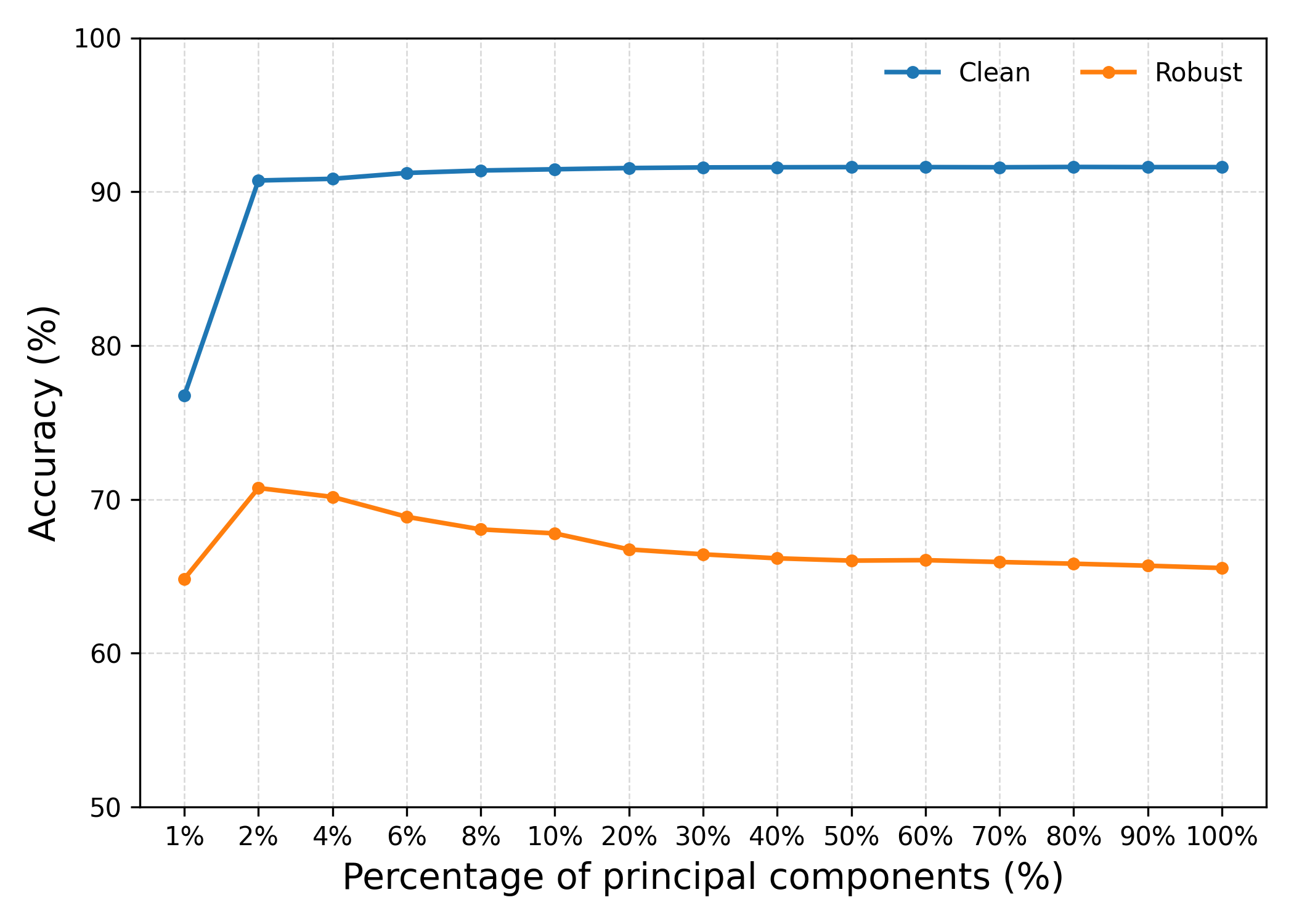}
    \caption{Clean and robust accuracy evaluated with a CIFAR-10 robust WRN-28-10 model when projecting features from natural or adversarial images onto the top-$K$ principal components before the classification head.}
    \label{fig:cls_dim_example}
\end{figure}

Figure~\ref{fig:cls_dim_example} illustrates the classification dimension analysis of Section~\ref{sec:distinct_roles} on a CIFAR-10 adversarially trained model.
Clean accuracy gradually recovers as more principal components are included, whereas robust accuracy first improves and then degrades as $K$ increases.

\section{SSL Features as Surrogate Representations in DRA}
\label{app:ssl_surrogate}

To examine whether the regularization benefit of DRA extends beyond diffusion representations, we replace the diffusion target with self-supervised features from DINOv3~\citep{simeoni2025dinov3}, MAE~\citep{he2022mae}, and CLIP~\citep{radford2021clip} in our ImageNet setup, keeping all other components of the AT recipe fixed.

In every configuration, training was unstable and collapsed within the first few epochs, with the AT loss failing to decrease. This suggests a strong conflict between the adversarial training objective and the SSL alignment loss.
We attempted to mitigate this by lowering the regularization strength from the default $\lambda{=}1.2$ to $\lambda \in \{0.6, 0.3, 0.1\}$, but the collapse persisted across all settings.

These results suggest that, while common SSL features can be effective as \emph{initialization} for robust finetuning (e.g., the DINOv3-initialized ImageNet experiments described in Section~\ref{sec:exp_setups}), they do not serve as useful \emph{regularization targets} in adversarial training.
This is likely because these methods lack the denoising-based generative pretraining procedure with noisy images,
and therefore do not learn representations that remain semantically stable and informative under input perturbations,
which is the critical property that makes diffusion features effective regularization targets in AT.
Nevertheless, DRA with diffusion representations incurs no additional pretraining cost, since diffusion models are already required for generating the synthetic data that has become critical in scaling AT.
Additionally, combining DRA with strong SSL-initialized backbones also leads to further improvements.

\section{Applicability of DRA with Foundation Diffusion Models}
\label{app:foundation_diffusion}

To examine whether DRA can benefit from an off-the-shelf diffusion teacher,
we use Stable Diffusion 1.5 (SD-1.5)~\citep{rombach2022ldm} as the alignment
target, and evaluate it using WRN-28-10 under the DM-AT setup with one million synthetic images.

\begin{table}[htbp]
    \centering
    \small
    \caption{Comparison of diffusion alignment targets on CIFAR-10 using
    WRN-28-10 and one million synthetic images. Accuracy is reported in
    percent. SD-1.5 is used without target-dataset finetuning.}
    \label{tab:foundation_diffusion}
    \begin{tabular}{lccc}
        \toprule
        Method & Clean & AA ($\ell_\infty=8/255$) & AA ($\ell_2=0.5$) \\
        \midrule
        DM-AT & 91.12 & 63.35 & 66.70 \\
        \quad +DRA, SD-1.5 & $91.59_{\pm.04}$ & $63.60_{\pm.13}$ & $68.73_{\pm.07}$ \\
        \quad +DRA, EDM & $\mathbf{92.32_{\pm.12}}$ & $\mathbf{64.09_{\pm.04}}$ & $\mathbf{69.19_{\pm.05}}$ \\
        \bottomrule
    \end{tabular}
\end{table}

Table~\ref{tab:foundation_diffusion} shows that SD-1.5 improves clean
accuracy and robust accuracy over the baseline DM-AT. 
These results demonstrate that DRA can benefit from an off-the-shelf
diffusion teacher without target-dataset finetuning, and using a diffusion model trained on the downstream dataset
could provide better performance.

\section{Noisy-Image Discriminative Pretrained Representations}

\label{app:noisy_disc_exp}
\begin{wrapfigure}{r}{0.6\textwidth}
    \vspace{-0.4cm}
    \centering
    \begin{subfigure}[t]{0.49\linewidth}
        \centering
        \includegraphics[height=3cm, keepaspectratio]{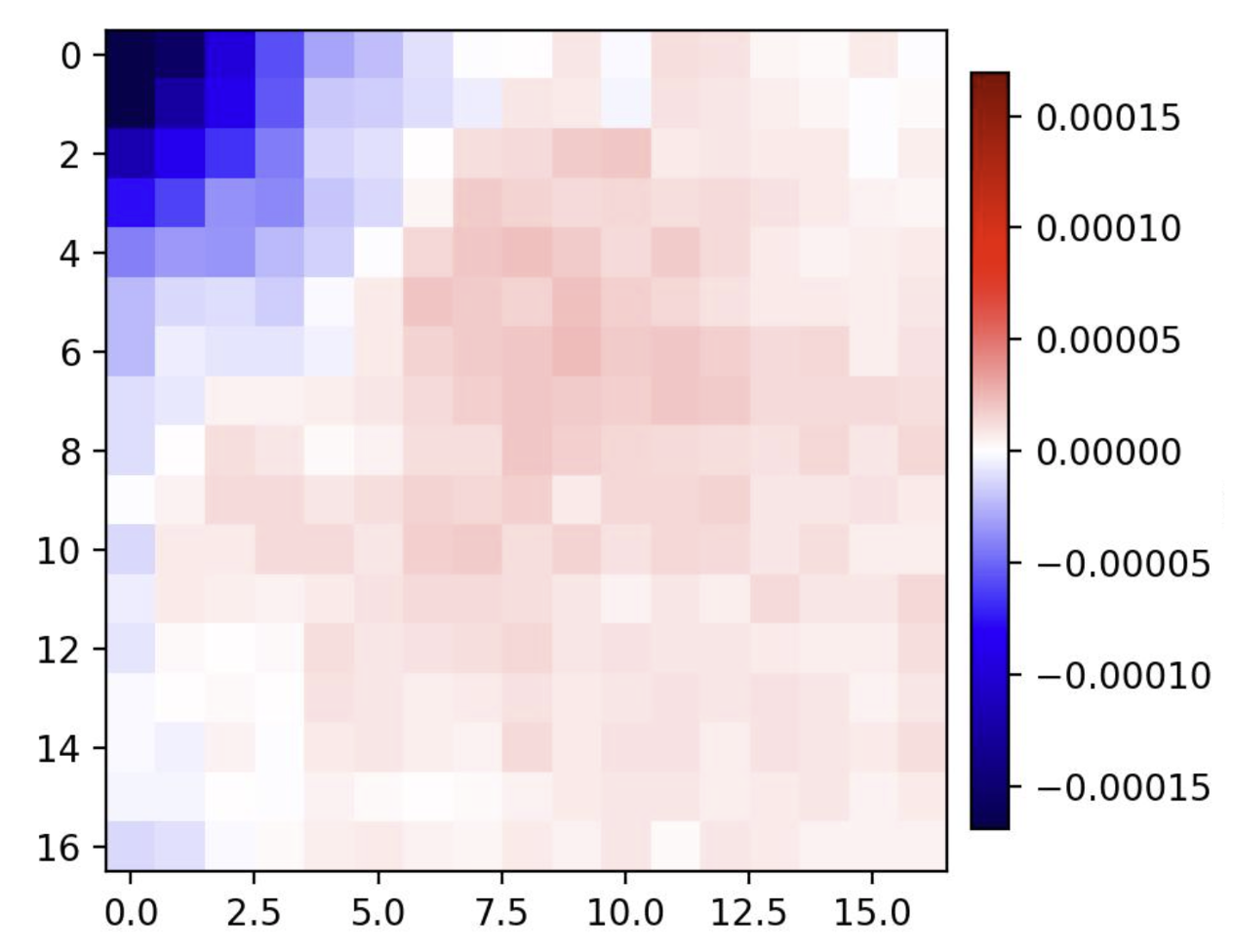}
        \caption{Noisy Discriminative}
        \label{fig:freq_diff_noisy}
    \end{subfigure}\hfill
    \begin{subfigure}[t]{0.49\linewidth}
        \centering
        \includegraphics[height=3cm, keepaspectratio]{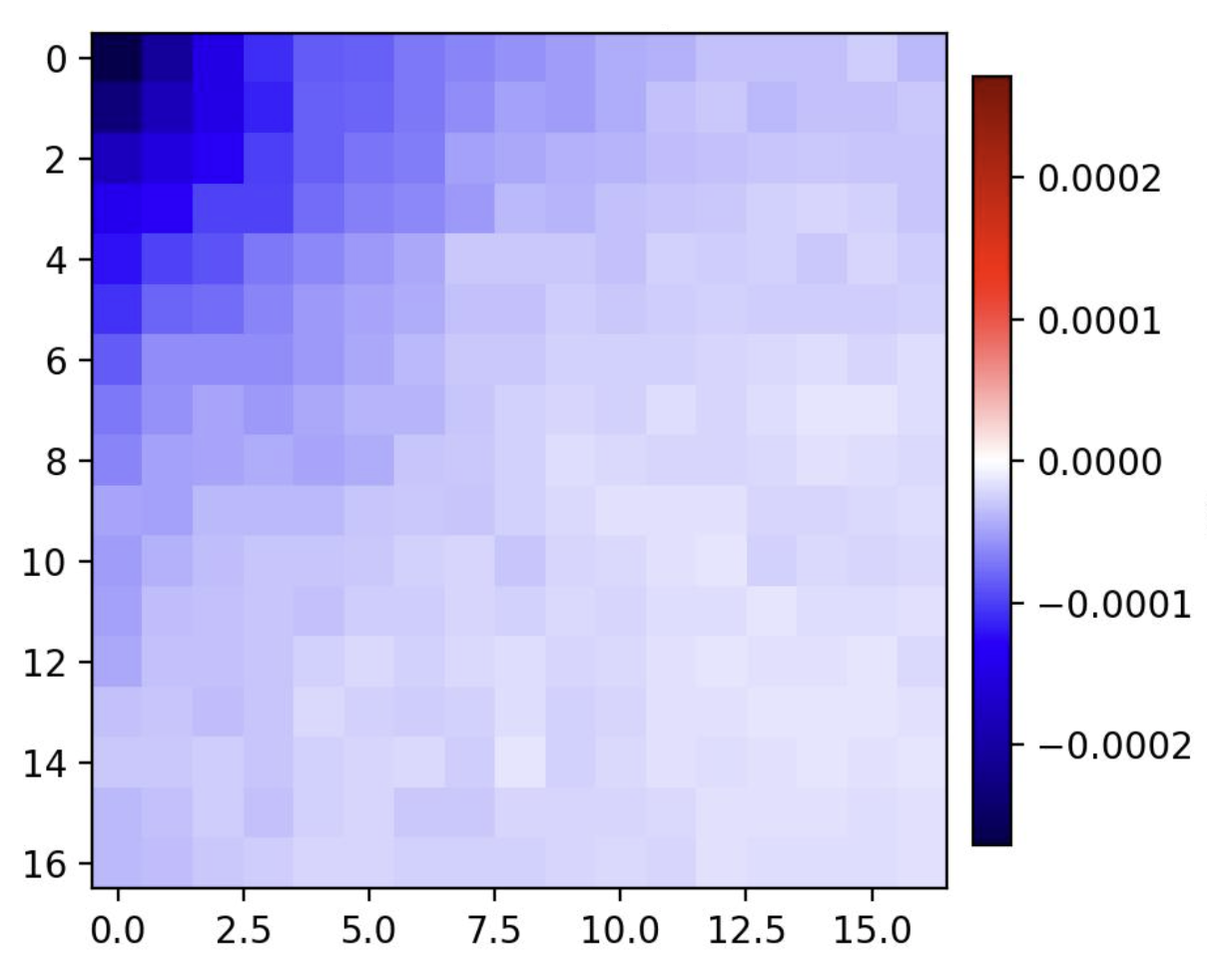}
        \caption{Diffusion Representations}
        \label{fig:freq_diff_dra}
    \end{subfigure}
    \caption{The input gradient frequency difference maps between DM-AT+DRA and DM-AT. (a) DRA aligned to noisy-input discriminative pretrained representations. (b) DRA aligned to the original diffusion representations. Positive values indicate increased sensitivity. Low-frequency components are in the upper-left.}
    \label{fig:freq_diff_map}
    \vspace{-0.3cm}
\end{wrapfigure}

We train a classifier with the same diffusion UNet encoder on the same diffusion noisy ($\sigma_t=0.1$) inputs, using cross-entropy loss.
Its accuracy is comparable to that of a linear probe trained on diffusion representations.
As discussed in Section~\ref{exp:ablation}, using these noisy-image discriminative representations as the alignment target degrades robustness.
To investigate the effect, we compute frequency maps on the input gradients and take the difference between DM-AT and DM-AT+DRA.
Figure~\ref{fig:freq_diff_map} shows that DRA with diffusion representations reduces the overall gradient magnitude, most strongly at low frequencies where sensitivity is concentrated, while alignment to noisy-image discriminative representations increases sensitivity to mid and high-frequency features.
Our results complement the findings of \citet{jaini2024intriguing}, who find that noisy discriminative training could lead to shape bias but decreased OOD classification performance.

\section{Ablation Study on the Sensitivity of Timesteps}
\label{app:timestep_sensitivity}

We examine DRA's sensitivity to the diffusion noise level used for feature
extraction. Using WRN-28-10 on CIFAR-10 under the DM-AT setup with one
million synthetic images, we vary $\sigma_t\in\{0.05,0.1,0.25,0.5\}$,
including the default value $\sigma_t=0.1$ described in
Appendix~\ref{app:DRA_Imp_Abl}.

\begin{table}[htbp]
    \centering
    \small
    \caption{Sensitivity to the diffusion noise level on CIFAR-10 using
    WRN-28-10 and one million synthetic images. Accuracy is reported in
    percent.}
    \label{tab:timestep_sensitivity}
    \begin{tabular}{lccc}
        \toprule
        Method & Clean & AA ($\ell_\infty=8/255$) & AA ($\ell_2=0.5$) \\
        \midrule
        DM-AT & 91.12 & 63.35 & 66.70 \\
        \quad +DRA ($\sigma_t=0.05$) & $\mathbf{92.48_{\pm.07}}$ & $\mathbf{64.30_{\pm.21}}$ & $68.65_{\pm.05}$ \\
        \quad +DRA ($\sigma_t=0.1$, default) & $92.32_{\pm.12}$ & $64.09_{\pm.04}$ & $\mathbf{69.19_{\pm.05}}$ \\
        \quad +DRA ($\sigma_t=0.25$) & $92.37_{\pm.16}$ & $64.07_{\pm.10}$ & $68.52_{\pm.13}$ \\
        \quad +DRA ($\sigma_t=0.5$) & $92.11_{\pm.13}$ & $63.91_{\pm.06}$ & $68.43_{\pm.12}$ \\
        \bottomrule
    \end{tabular}
\end{table}

As shown in Table~\ref{tab:timestep_sensitivity}, all tested noise levels
improve clean accuracy and robust accuracy over the DM-AT baseline.
The results show that the improvement over the baseline is robust around the current default timestep.

\newpage

\section{Additional Sweep of Regularization Strength}
\label{app:reg_sweep}

\begin{wrapfigure}{r}{0.40\textwidth}
    \vspace{-0.4cm}
    \centering
    \includegraphics[width=\linewidth]{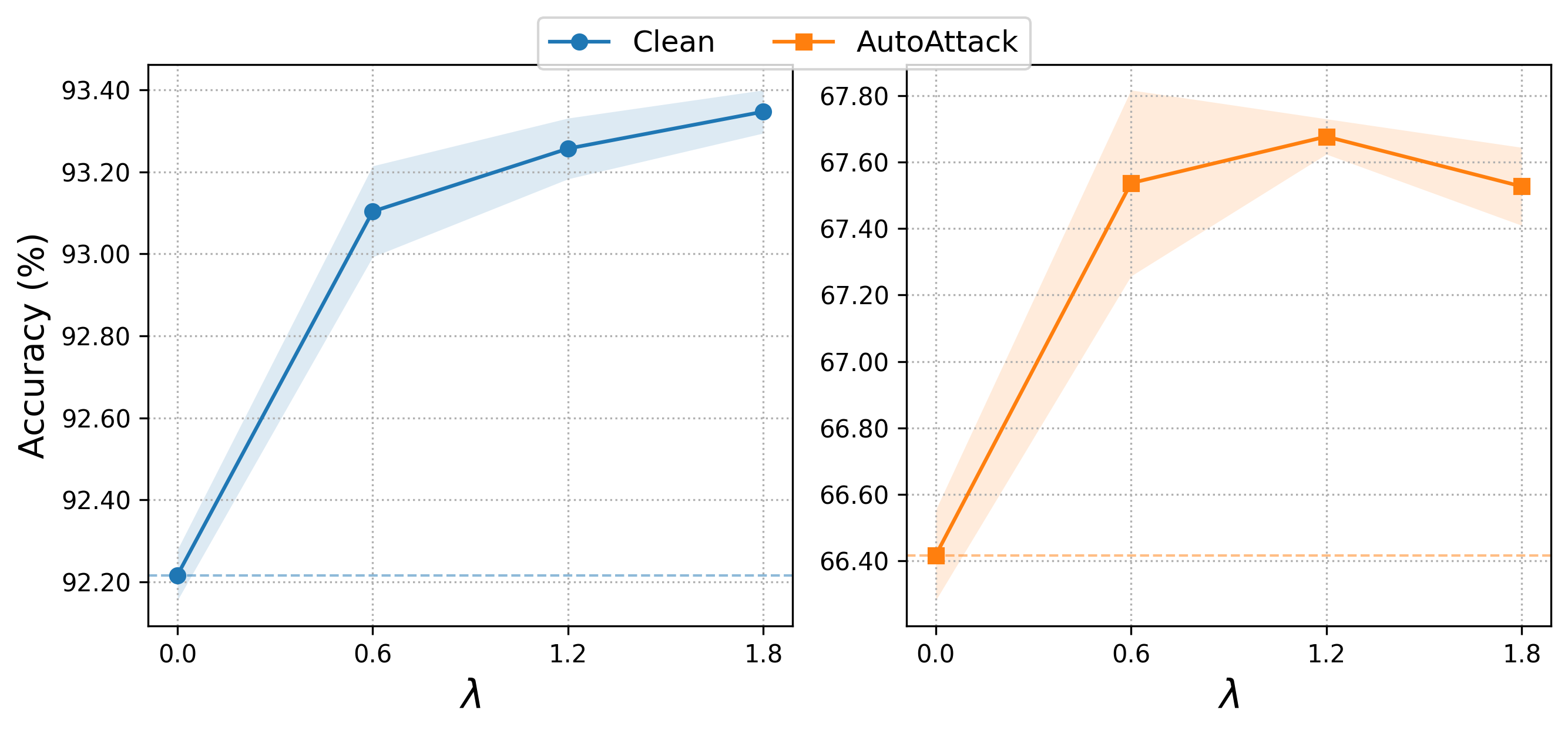}
    \caption{Sweep of the regularization strength $\lambda$ on CIFAR-10 with 20M synthetic images using ViT-B/2.}
    \label{fig:vit_reg_sweep}
    \vspace{-0.5cm}
\end{wrapfigure}

In addition to Figure~\ref{fig:reg_strength}, we conduct an additional sweep of $\lambda \in \{0.0, 0.6, 1.2, 1.8\}$ for ViT-B/2 on CIFAR-10 with 20M synthetic images.
Figure~\ref{fig:vit_reg_sweep} shows that any non-zero $\lambda$ improves both clean and AutoAttack robust accuracy over the DM-AT baseline, with robust accuracy peaking at $\lambda{=}1.2$.
Due to the cost of running adversarial training, we did not sweep $\lambda$ for every other architecture and synthetic-data scale; the purpose here is to show that a fixed $\lambda{=}1.2$ is sufficient to demonstrate the effectiveness of DRA. In practice, a small sweep of $\lambda$ may further improve results in a given setting.

\section{Additional ImageNet Experiments}
\label{app:additional_in1k_exp}

\begin{wraptable}{r}{0.45\textwidth}
    \vspace{-0.5cm}
    \centering
    \caption{ImageNet results for ViT-B/16 trained from scratch and from a DINOv3-initialized checkpoint.
    AutoAttack is evaluated at $\ell_\infty=4/255$.}
    \label{tab:in1k_scratch}
    \begin{tabular}{lcc}
        \toprule
        Model & Clean & AA  \\
        \midrule
        ViT-B/16 (from scratch) & 70.69 & 48.06 \\
        \quad +DRA             & \textbf{73.77} & \textbf{50.24} \\
        \midrule
        ViT-B/16 (DINOv3) & 74.67 & 54.88 \\
        \quad +DRA             & \textbf{76.86} & \textbf{55.53} \\
        \bottomrule
    \end{tabular}
    \vspace{-0.3cm}
\end{wraptable}

The ImageNet experiments in Table~\ref{tab_imagenet} finetune from a DINOv3 checkpoint to demonstrate the effectiveness of DRA in real-world scenarios when strong pretrained features are available.
To examine whether DRA also helps without such an initialization, we additionally train ViT-B/16 from scratch on ImageNet.
Table~\ref{tab:in1k_scratch} reports the results.
DRA improves both clean and robust accuracy in both regimes.
This indicates that DRA is more effective when there is no strong pretrained feature initialization, while still providing additional benefits when strong pretrained representations are available.

\section{Limitations}
\label{app:limitations}

The contribution of our work focuses on a new perspective on diffusion models through representation usage in adversarial training, 
viewing them as a source of diverse yet partially robust feature priors in addition to synthetic data generators.
Hence, the broader empirical questions of which properties of a teacher representation make it suitable as a regularization target in adversarial training,
and also what other methods could more effectively extract robust knowledge out of diffusion models to build robust discriminative classifiers,
are promising directions that require future investigation.
Additionally, our experiments focus on the standard $\ell_\infty$ adversarial robustness for image classification on CIFAR-10/100 and ImageNet;
extending DRA to other modalities, tasks, or threat models for robust visual recognition is a natural direction for future work.

\section{Broader Impacts}
\label{app:broader_impact}

Adversarial robustness contributes to safer deployment of machine learning in security-sensitive applications such as content moderation, biometric verification, and medical imaging,
and our work makes diffusion-based adversarial training recipes more effective by reusing diffusion features as an auxiliary learning signal, without additional data or substantial extra compute.
On the negative side, adversarial training is computationally expensive and contributes to the energy footprint of model development;
robustness measured on standard benchmarks (e.g., AutoAttack on $\ell_\infty$ perturbations) does not guarantee robustness under threat models that are not included in the test scenarios and may give a false sense of security.

\section{Reproducibility}
\label{app:reproducibility}
To ensure the reproducibility of our experiments, we have provided detailed implementation of the proposed method.
Additionally, the code implementation is available at \url{https://github.com/pinhank0121/DRA}, enabling readers to replicate the experiments.

\section{Licenses for Existing Assets and LLM Usage Declaration}
\label{app:asset_licenses}
Table~\ref{tab:asset_licenses} lists the existing assets used in this work,
including datasets, pretrained models, and code packages, together with their licenses.
We use all assets in accordance with their respective licenses for non-commercial academic research.

\textbf{LLM assistance.} LLM tools were used for assisting code implementations and manuscript editing.

\begin{table}[h]
\centering
\small
\caption{Existing assets used in this work and their licenses.}
\label{tab:asset_licenses}
\scalebox{0.9}{
\begin{tabular}{llll}
\toprule
Asset & Type & Reference & License \\
\midrule
CIFAR-10 / CIFAR-100 & Dataset & \citet{krizhevsky2009learning} & Custom (research use) \\
ImageNet (ILSVRC-2012) & Dataset & \citet{russakovsky2015imagenet} & Custom (research use) \\
DM-AT & Dataset / Code & \citet{wang2023better} & MIT \\
EDM diffusion model & Model / Code & \citet{karras2022edm} & CC BY-NC-SA 4.0 \\
LightningDiT & Model / Code & \citet{yao2025ldit} & MIT \\
DINOv3 pretrained checkpoints & Model & \citet{simeoni2025dinov3} & DINOv3 License \\
\texttt{timm} library & Model / Code & \citet{rw2019timm} & Apache 2.0 \\
RobustBench & Model / Code & \citet{croce2021robustbench} & MIT \\
AutoAttack & Code & \citet{croce2020reliable} & MIT \\
\bottomrule
\end{tabular}
}
\end{table}

\clearpage
\section*{NeurIPS Paper Checklist}

\begin{enumerate}

\item {\bf Claims}
    \item[] Question: Do the main claims made in the abstract and introduction accurately reflect the paper's contributions and scope?
    \item[] Answer: \answerYes{}
    \item[] Justification: The main claims made in the abstract and introduction accurately reflect the paper's contributions and scope.

\item {\bf Limitations}
    \item[] Question: Does the paper discuss the limitations of the work performed by the authors?
    \item[] Answer: \answerYes{}
    \item[] Justification: The paper discusses the limitations of the work in the Appendix. 

\item {\bf Theory assumptions and proofs}
    \item[] Question: For each theoretical result, does the paper provide the full set of assumptions and a complete (and correct) proof?
    \item[] Answer: \answerNA{}
    \item[] Justification: The paper does not contain theoretical results.

    \item {\bf Experimental result reproducibility}
    \item[] Question: Does the paper fully disclose all the information needed to reproduce the main experimental results of the paper to the extent that it affects the main claims and/or conclusions of the paper (regardless of whether the code and data are provided or not)?
    \item[] Answer: \answerYes{}
    \item[] Justification: Implementation details are provided in the appendix, covering computational resources, experimental settings, hyperparameters.

\item {\bf Open access to data and code}
    \item[] Question: Does the paper provide open access to the data and code, with sufficient instructions to faithfully reproduce the main experimental results, as described in supplemental material?
    \item[] Answer: \answerYes{}
    \item[] Justification: Sufficient instructions are provided in the appendix, and the code implementation is provided.

\item {\bf Experimental setting/details}
    \item[] Question: Does the paper specify all the training and test details (e.g., data splits, hyperparameters, how they were chosen, type of optimizer) necessary to understand the results?
    \item[] Answer: \answerYes{}
    \item[] Justification: The training and test details, including hyperparameters and settings, are provided with sufficient instructions in the appendix.

\item {\bf Experiment statistical significance}
    \item[] Question: Does the paper report error bars suitably and correctly defined or other appropriate information about the statistical significance of the experiments?
    \item[] Answer: \answerYes{}
    \item[] Justification: The mean and standard deviation over three runs are reported for the main experiments in the paper.

\item {\bf Experiments compute resources}
    \item[] Question: For each experiment, does the paper provide sufficient information on the computer resources (type of compute workers, memory, time of execution) needed to reproduce the experiments?
    \item[] Answer: \answerYes{}
    \item[] Justification: The paper provides sufficient information on the computer resources needed to reproduce the experiments in the appendix.
    
\item {\bf Code of ethics}
    \item[] Question: Does the research conducted in the paper conform, in every respect, with the NeurIPS Code of Ethics \url{https://neurips.cc/public/EthicsGuidelines}?
    \item[] Answer: \answerYes{}
    \item[] Justification: This paper conforms with the NeurIPS Code of Ethics.

\item {\bf Broader impacts}
    \item[] Question: Does the paper discuss both potential positive societal impacts and negative societal impacts of the work performed?
    \item[] Answer: \answerYes{}
    \item[] Justification: The paper discusses both potential positive societal impacts and negative societal impacts of the work performed in the Broader Impacts section of the appendix.
    
\item {\bf Safeguards}
    \item[] Question: Does the paper describe safeguards that have been put in place for responsible release of data or models that have a high risk for misuse (e.g., pre-trained language models, image generators, or scraped datasets)?
    \item[] Answer: \answerNA{}
    \item[] Justification: The paper poses no such risks.

\item {\bf Licenses for existing assets}
    \item[] Question: Are the creators or original owners of assets (e.g., code, data, models), used in the paper, properly credited and are the license and terms of use explicitly mentioned and properly respected?
    \item[] Answer: \answerYes{}
    \item[] Justification: All datasets, pretrained models, and code packages used in this work are publicly available existing assets. We cite the original papers and use all assets in accordance with their respective licenses and terms of use. A complete list of assets, with their sources and licenses, is provided in Appendix~\ref{app:asset_licenses}.

\item {\bf New assets}
    \item[] Question: Are new assets introduced in the paper well documented and is the documentation provided alongside the assets?
    \item[] Answer: \answerNA{}
    \item[] Justification: The paper does not introduce new assets.

\item {\bf Crowdsourcing and research with human subjects}
    \item[] Question: For crowdsourcing experiments and research with human subjects, does the paper include the full text of instructions given to participants and screenshots, if applicable, as well as details about compensation (if any)? 
    \item[] Answer: \answerNA{}
    \item[] Justification: The paper does not involve crowdsourcing nor research with human subjects.

\item {\bf Institutional review board (IRB) approvals or equivalent for research with human subjects}
    \item[] Question: Does the paper describe potential risks incurred by study participants, whether such risks were disclosed to the subjects, and whether Institutional Review Board (IRB) approvals (or an equivalent approval/review based on the requirements of your country or institution) were obtained?
    \item[] Answer: \answerNA{}
    \item[] Justification: The paper does not involve research with human subjects.

\item {\bf Declaration of LLM usage}
    \item[] Question: Does the paper describe the usage of LLMs if it is an important, original, or non-standard component of the core methods in this research? Note that if the LLM is used only for writing, editing, or formatting purposes and does \emph{not} impact the core methodology, scientific rigor, or originality of the research, declaration is not required.
    \item[] Answer: \answerNA{}
    \item[] Justification: LLM tools were used for assisting code implementations and manuscript editing.

\end{enumerate}

\end{document}